\documentclass[a4paper,11pt]{article}
\usepackage[utf8]{inputenc}
\usepackage[T1]{fontenc}
\usepackage[french,english]{babel}
\usepackage[margin=2.8cm]{geometry}

\usepackage{courier}
\usepackage{amsmath,amsfonts,amssymb,amsthm}
\usepackage{graphicx}
\usepackage{natbib}
\usepackage{color}
\usepackage{comment}
\usepackage{hyperref}
\usepackage{rotating}
\usepackage{authblk}

\newcommand{\Rlogo}{\protect\includegraphics[height=1.8ex,keepaspectratio]{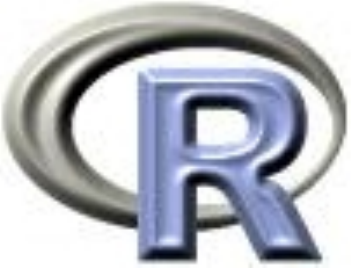}}

\newcommand*{\E}{\mathbb{E}}
\newcommand*{\EE}[1]{\E\left[#1\right]}

\newcommand*{\R}{\mathbb{R}}

\newcommand*{\nrm}[1]{\left\| #1 \right\|}            % norme
\newtheorem{thm}{Theorem}[section]

\newtheorem{lem}[thm]{Lemma}

\title{Online Principal Component Analysis in High Dimension:\\ Which Algorithm to Choose?}

\author{Herv\'e \textsc{Cardot}$^{(\textrm{a})}$ and David \textsc{Degras}$^{(\textrm{b})}$ \\
(a) Institut de Math\'ematiques de Bourgogne, \\ Universit\'e de Bourgogne Franche Comt\'e \\
(b) DePaul University }
\begin{document}

\maketitle

%\tableofcontents
\begin{abstract}
In the current context of data explosion, online techniques that do not require storing all data in memory are indispensable to routinely perform tasks like principal component analysis (PCA). Recursive algorithms that update the PCA  with each new observation have been studied in various fields of research  
%(e.g.,  statistics, numerical analysis, signal processing) 
and found wide applications in industrial  monitoring, computer vision, astronomy, and latent semantic indexing, among others. This work provides guidance for selecting an online PCA algorithm in practice. We present the main approaches to online PCA, namely, perturbation techniques, incremental methods, and stochastic optimization, and compare their statistical accuracy, computation time, and memory requirements using artificial and real data.  
Extensions to missing data and to functional data are discussed. 
%An application of online PCA to face recognition is presented. 
%In this setup we propose an efficient bootstrap procedure for selecting smoothing parameters. 
All studied algorithms are available in the  \Rlogo~package \texttt{onlinePCA} on CRAN.
\end{abstract}

Incremental SVD,  
%Krasulina's algorithm, Oja's algorithm, 
Perturbation methods,  Recursive algorithms, Stochastic gradient.

%%%%%%%%%%%%%%%%%%%%%%%%%%%%%%%%%%%%%%%%%%%%%%%%%%%%%%%%%%%%%%%

\section{Introduction}
\label{sec:intro}

Principal Component Analysis (PCA) is a popular method for reducing the dimension of data while preserving most of their variations (see \cite{Jolliffe2002} for a general presentation).  
In a few words, PCA consists in extracting the main modes of variation of the data around their mean via the computation of new synthetic variables named  principal components. 
This technique has found applications in fields as diverse as data mining, industrial process modeling \citep{Tang2012}, face recognition \citep{Zhao2006}, latent semantic indexing \citep{ZhaSimon1999}, sentiment analysis \citep{IodiceDEnzaMarkos2015}, astronomy \citep{Budavarietal2009}, and more.
Traditional PCA, also called {\it batch} PCA or {\it offline} PCA, is typically implemented via the eigenvalue decomposition (EVD) of the sample covariance matrix or the singular value decomposition (SVD) of the centered data.  EVD and SVD can be standardly computed in $O(nd \min(n,d))$ floating point operations (flops), where $n$ is the sample size and $d$ the number of variables \citep[][Chap. 8]{GolubVanLoan2013}. % GolubReinsch70
If the number $q$ of required principal components is much smaller than $n$ and $d$ (this is usually the case), approximate PCA decompositions can be obtained quickly with little loss of accuracy. For example, truncated SVD can be computed in $O(ndq)$ flops by combining rank-revealing QR factorization and R-SVD. Other effective approaches to  truncated SVD or EVD include power methods, Krylov subspace methods \citep{Lehoucqetal1998,GolubVanLoan2013}, and random projection methods \citep{HMT2011}. 
With respect to memory usage, batch PCA algorithms have at least $O(nd)$ space complexity as they necessitate to 
 hold all data in computer memory. EVD requires $O(d^2)$ additional memory for storing a  $d\times d$ 
covariance matrice (or $O(n^2)$ memory to store a $n\times n$ covariance matrix, see Section 2). 
Because of its time and space complexity, batch PCA is essentially infeasible with: (i) massive datasets for which $n$ and $d$ are, say, in the thousands or millions, and (ii) datasets that change rapidly and may need to be processed on the fly (e.g., streaming data, databases). Given the exponential growth of such data in modern applications, fast and accurate PCA algorithms are in high demand. 

Over the years a large number of solutions has emerged from fields as diverse as signal processing, statistics, numerical analysis, and machine learning. These approaches, called {\it recursive, incremental}, or {\it online} PCA in the literature, consist in updating the current PCA each time new data are observed without recomputing it from scratch. 
This updating idea does not only apply to time-varying datasets but also to datasets that are too large to be processed as a whole and must be analyzed one subset at a time. We give here a brief description of the main approaches to online PCA. Some of the most representative techniques will be examined in detail in the next sections. 
One of the first historical approaches is the numerical resolution of so-called {\it secular equations}  \citep{Golub73,GuEisenstat94,Lietal2000}. By exploiting the interlacing property of the eigenvalues of a covariance matrix under rank 1 modifications,  this approach reduces the PCA update to finding the roots of a rational function. Interestingly, this recursive method is exact, that is, it produces the same results as batch PCA.  
{\it Perturbation methods} formulate the PCA update as the EVD of a diagonal matrix plus a rank-1 matrix and obtain closed-form solutions using large-sample approximations \citep{Hegdeetal2006}. 
{\it Incremental SVD} is a highly effective method for producing an approximate, reduced rank PCA. It allows for block updates and typically involves the SVD or EVD of a small matrix of dimension $(q+r)\times (q+r)$, where $r$ is the block size. Important contributions in this area include \cite{ZhaSimon1999}, \cite{Brand02incrementalsingular}, and \cite{Bakeretal2012}.
{\it Stochastic optimization} methods enable very fast PCA updates and bear interesting connections with neural networks.  Prominent examples are the stochastic gradient algorithms of \cite{Krasulina1970}, \cite{OjaKarhunen1985}, \cite{Oja1992}, and the generalized hebbian algorithm of \cite{Sanger1989}. While many results can be found on the consistency of these methods, their finite-sample behavior does not lend itself well to scrutiny. In addition, their numerical performances hinge on tuning parameters that are often selected by trial-and-error. Related techniques that are less sensitive to the choice of tuning parameters include \cite{WZH2003} and \cite{Mitliagkasetal2014}. 
Approaches like {\it moving window PCA} \citep{Wang2005} and {\it fixed point algorithms} (\cite{RaoPrincipe2000}) have also garnered interest. 
In recent years, {\it randomized} algorithms for online PCA have been developed in computer science and machine learning  \citep{WK2008,BoutsidisGKL15}. An important aspect of this line of research is to establish bounds on finite-sample performance. 

In contrast to the wide availability of online PCA methods, very few studies provide guidance on selecting a method in practice. To our knowledge, the only articles on this topic are \cite{Chatterjee2005}, 
%compares ten stochastic algorithms in the estimation of the first principal eigenvector using low-dimensional data;  
\cite{Aroraetal2012}, and \cite{Ratoetal2015}. This  gap in the literature is a significant problem because each application has its own set of data, analytic goals, computational resources, and time constraints, and no single online PCA algorithm can perform satisfactorily in all situations, let alone optimally. 
%  (data size, analytic goals, time/space constraints, missingness, {\it etc}). 
The present paper considerably expands on previous work in terms of scope of study and practical recommendations.    
We investigate a wide range of popular online PCA algorithms and  compare their computational cost (space and time complexity, computation time, memory usage) and statistical accuracy in various setups: simulations and real data analyses, low- and high-dimensional data. We discuss practical issues such as the selection of tuning parameters, the potential loss of orthogonality among principal components, and 
 the choice of vector versus block updates. Particular attention is given to the analysis of functional data and to the imputation of missing data. All algorithms under study are implemented in the R package \texttt{onlinePCA} available at \url{http://cran.r-project.org/package=onlinePCA}.

The remainder of the article is organized as follows: Section \ref{sec: batch pca} provides a reminder on batch PCA.   
Section \ref{sec:perturbation} gives recursive formulae for the sample mean and covariance.  
It also presents approximate and exact perturbation methods for online PCA. 
Incremental PCA is discussed in Section \ref{sec:incremental} and stochastic approximation methods are detailed in Section \ref{sec: stochastic}. 
Extensions to nonstationary data, missing values, and functional data are given in Section \ref{sec: extensions}. 
In Section \ref{sec: simulations}, the computational and statistical performances of the previous online PCA methods
are compared in simulations and in a face recognition application. Concluding remarks are offered in Section \ref{sec: conclusion}. R code for the numerical study of Section \ref{sec: simulations} 
is available online as Supplementary Materials.

\section{Batch PCA}
\label{sec: batch pca}

% Suppose we have
Given data vectors $\mathbf{x}_1,\ldots,\mathbf{x}_n$  in $\R^d$, with $d$ possibly large, 
batch PCA seeks to produce a faithful  representation of the data 
in a low-dimensional subspace of $\R^d$. 
%subspace of $\R^d$ that faithfully represents the data. 
% that we want to represent in a low-dimensional subspace of $\R^d$, . The dimension $q$ of the subspace should be small while not losing much information, the loss being measured by the mean squared norm between the (centered) initial vectors and their projections in the subspace. 
The dimension $q$ of the subspace must be small enough to effectively reduce the data dimension, 
but large enough to retain most of the data variations. 
The quality of the representation is measured by the  squared distance 
between the (centered) vectors and their projections in the subspace. 
Denoting the sample mean by 
$\boldsymbol{\mu}_n = \frac{1}{n} \sum_{i=1}^n \mathbf{x}_i$, the goal of batch PCA 
is thus to find a projection matrix $\mathbf{P}_q$ of rank $q\le d$  that minimizes the loss function 
\begin{align}
R_n(\mathbf{P}_q) &= \frac{1}{n} \sum_{i=1}^n \big\| \left( \mathbf{x}_i  -\boldsymbol{\mu}_n \right) - \mathbf{P}_q \left( \mathbf{x}_i  -\boldsymbol{\mu}_n \right) \big\|^2.
\label{def:riskPCA}
\end{align} 
%is minimum. 

Consider the sample covariance matrix
\begin{align}
\boldsymbol{\Gamma}_n & = \frac{1}{n} \sum_{i=1}^n \left( \mathbf{x}_i - \boldsymbol{\mu}_n \right) \left( \mathbf{x}_i - \boldsymbol{\mu}_n \right)^T .
\label{def:Gammaest}
\end{align}
Let $\mathbf{u}_{1,n}, \ldots, \mathbf{u}_{d,n}$ be orthonormal eigenvectors of   $\boldsymbol{\Gamma}_n$ with associated  eigenvalues $\lambda_{1,n} \geq \ldots \geq \lambda_{d,n}  \geq 0$.
The minimum of $R_n$ among rank $q$ projection matrices 
is attained when $\mathbf{P}_q$ is the orthogonal projector 
$ \sum_{j=1}^q \mathbf{u}_{j,n}\mathbf{u}_{j,n}^T$, in which case $R_n(\mathbf{P}_q) = \sum_{j=q+1}^d \lambda_{j,n}$.
In other words, batch PCA reduces to finding the first $q$ eigenvectors of the covariance $\boldsymbol{\Gamma}_n$ or, equivalently, the first $q$ singular vectors of $\mathbf{X}$. These eigenvectors are called principal components and from here onwards, we use the two expressions interchangeably. 

%\begin{rem}
%PCA can also be based on $(n-1)^{-1}\sum_i (\mathbf{x}_i - \boldsymbol{\mu}_n ) ( \mathbf{x}_i - \boldsymbol{\mu}_n )^T$, the adjusted empirical covariance matrix. This produces the same eigenvectors (variable loadings) and principal components but  slightly different eigenvalues (multiplicative factor $n/(n-1)$).
%In a probabilistic setting where the $\mathbf{x}_i$ are independent realizations of a random vector $\mathbf{X}$ with mean $\mu$ and covariance $\boldsymbol{\Gamma}$, Jensen's inequality shows that on average, the first eigenvalue of the adjusted empirical covariance slightly overestimates the first eigenvalue of $\boldsymbol{\Gamma}$ (positive bias). 
%In this context, the factor $1/n$ of $\boldsymbol{\Gamma}_n$ provides an estimator with lower mean squared error.
%\end{rem}

The computation of $\boldsymbol{\Gamma}_n$ takes $O(nd^2)$ flops and its full EVD requires $O(d^3)$ flops.
When $d$ is large and $q \ll d$, iterative methods such as the power method and implicitly restarted methods 
should be preferred to standard EVD as they are much faster and approximate the first $q$ eigenvectors of $\boldsymbol{\Gamma}_n$ with high accuracy \citep[\textit{e.g.,}][]{Lehoucqetal1998,GolubVanLoan2013}. 
Despite this speedup, the  cost $O(nd^2)$ of computing $\boldsymbol{\Gamma}_n$ does not scale with the data. 
Note that  if $n\ll d $, batch PCA can be performed on $\mathbf{X}^T$, leading to the same result with reduced time complexity $O(n^2d + n^3)=O(n^2d)$ and space complexity $O(nd + n^2)=O(nd)$.
The duality between the PCA of $\mathbf{X}$ and that of $\mathbf{X}^T$ comes from the well-known result that $\mathbf{X}^T \mathbf{X}$ and $\mathbf{XX}^T$ have the same eigenvalues 
and that their sets of eigenvectors can be deduced from one another by left-multiplication by $\mathbf{X}$ 
or $\mathbf{X}^T$; see \cite{Holmes2008} for more details. In terms of space complexity, batch PCA necessitates having both $\mathbf{X}$ and $\boldsymbol{\Gamma}_n$ in random access memory, which incurs a storage cost of  $O(nd + d^2)=O(d^2)$.

When data arrive sequentially, performing a batch PCA for each new observation may not be feasible for several reasons: 
first, as $n$ and/or $d$ becomes large, the runtime $O(nd \min(n,d))$ of the algorithm becomes excessive and forbids   processing data on the fly; second, batch PCA requires storing all data, which is not always possible.    
Under these circumstances, much faster recursive techniques are of great interest. 
%that allow automatic update of the eigenvectors and the eigenvalues 
The price to pay for computational efficiency is that the obtained solution is often only an approximation to the true eigenelements.

%With a different point of view, non recursive but fast techniques have also been studied recently for fast covariance estimation when the data are functions (see  \cite{XRZC2014}).

%%%%%%%%%%%%%%%%%%%%%%%%%%%%%%%%%%%%%%%%%%%%%%%%%%%%%%%%%%%%%%%

%\section{Approximations based on perturbations of the covariance matrix}
\section{Perturbation methods}
\label{sec:perturbation}

%\subsection*{Recursive computation of the empirical mean and covariance}
%\label{sec: recursive mean and covariance}

The sample mean vector can be computed recursively as 
\begin{equation}\label{def:recurMean1}
\boldsymbol{\mu}_{n+1} =  \frac{n}{n+1}\,  \boldsymbol{\mu}_{n} + \frac{1}{n+1} \,\mathbf{x}_{n+1} 
\end{equation}
and similarly,  the sample covariance matrix satisfies   
% VERSION CENTREE SUR MU_N
\begin{equation}\label{def:recurGamma1}
\boldsymbol{\Gamma}_{n+1} = \frac{n}{n+1} \boldsymbol{\Gamma}_{n} +  \frac{n}{(n+1)^2} \left(\mathbf{x}_{n+1} - \boldsymbol{\mu}_{n} \right)\left(\mathbf{x}_{n+1} - \boldsymbol{\mu}_{n} \right)^T  .
\end{equation}
Note that for each new observation, the sample covariance matrix is updated through a rank one modification. 
In view of  \eqref{def:recurGamma1}, $\boldsymbol{\Gamma}_{n+1} $ can be expressed as a perturbation of $\boldsymbol{\Gamma}_{n} $:
\begin{equation}\label{def:recurGamma-as}
 \boldsymbol{\Gamma}_{n+1} = \boldsymbol{\Gamma}_{n} - \frac{1}{n+1} \left(  \boldsymbol{\Gamma}_{n} - \frac{n}{n+1} \left(\mathbf{x}_{n+1} - \boldsymbol{\mu}_{n} \right)\left(\mathbf{x}_{n+1} - \boldsymbol{\mu}_{n} \right)^T  \right). 
  \end{equation}
Hence, if the eigendecomposition of $\boldsymbol{\Gamma}_n$ is known, a natural idea to compute the eigenelements of $\boldsymbol{\Gamma}_{n+1}$ is to apply perturbation techniques to $\boldsymbol{\Gamma}_n$, provided that $n$ is sufficiently large and $n^{-1} \left(\mathbf{x}_{n+1} - \boldsymbol{\mu}_{n} \right)\left(\mathbf{x}_{n+1} - \boldsymbol{\mu}_{n} \right)^T$ is small compared to $\boldsymbol{\Gamma}_n$ 
(viewing $\mathbf{x}_1,\ldots, \mathbf{x}_n$ as independent copies of a random vector with finite variance), the latter condition is satisfied with probability tending to one as $n \to \infty$).   
Note that the trace of $\boldsymbol{\Gamma}_n$, 
which represents the total variance of the data, can also be computed recursively thanks to \eqref{def:recurGamma1}.

% \begin{rem}
%If the population mean $\boldsymbol{\mu}$ is known, then $\boldsymbol{\mu}_n$ should be replaced by $\boldsymbol{\mu}$ in the covariance expression (\ref{def:Gammaest})
%and the recursion formula \eqref{def:recurGamma1} becomes
%\begin{align}
%\boldsymbol{\Gamma}_{n+1} &= \frac{n}{n+1} \boldsymbol{\Gamma}_{n} +  \frac{1}{n+1}  \left(\mathbf{x}_{n+1} - \boldsymbol{\mu} \right)\left(\mathbf{x}_{n+1} - \boldsymbol{\mu} \right)^T.  
%\label{def:muknown}
%\end{align}
%\end{rem}

%\begin{rem}
%The trace of the empirical covariance matrix, 
%which represents the total variance of the data, 
% can  be computed recursively as 
%%in a very simple way.  We have
%\begin{align}
%\mathrm{tr}
%\left(\boldsymbol{\Gamma}_{n+1}\right) &= \frac{n}{n+1} \mathrm{tr}\left(\boldsymbol{\Gamma}_{n}\right) +  \frac{1}{n+1}  \left\|\mathbf{x}_{n+1} - \boldsymbol{\mu}_n \right\|^2.  
%\label{def:tracemuknown}
%\end{align}
%%It is thus asy to estimate recursively the total variance of the data, which is the sum of the eigenvalues of the covariance matrix.
%\end{rem}

\subsection*{Large-sample approximations}

We first recall a classical result in perturbation theory  of linear operators (see \cite{Kato1976} and \cite{Sibson1979} for a statistical point of view). Denote by $\mathbf{A}^+$ the Moore-Penrose pseudoinverse of a matrix $\mathbf{A}$. 
\begin{lem}\label{lem:perturb}
Let $\mathbf{B}$ be a symmetric matrix with eigenelements $(\lambda_j,\mathbf{v}_j), \hspace*{0.2mm} j=1,\ldots,d$. 
\hspace*{-0.7mm}Consider the first-order perturbation 
% that is perturbated to (assuming that  enough)
 \begin{align*}
 \mathbf{B}(\delta) &= \mathbf{B} + \delta \mathbf{C} + O(\delta^2)
 \end{align*}
 with $\delta$ small and $\mathbf{C}$ a symmetric matrix.
Assuming that all eigenvalues of $\mathbf{B}$ are distinct, 
the eigenelements of $ \mathbf{B}(\delta) $ satisfy 
\begin{align*}
\lambda_j(\delta) & = \lambda_j + \delta\, \langle \mathbf{v}_j, \mathbf{C v}_j\rangle + O(\delta^2) ,\\
\mathbf{v}_j(\delta) &= \mathbf{v}_j + \delta \, (\lambda_j \mathbf{I} - \mathbf{B})^+ %{-1}
\mathbf{Cv}_j + O(\delta^2).
\end{align*}
\end{lem}

% VERSION CENTERED ON MU_N
We now apply Lemma \ref{lem:perturb} 
to  (\ref{def:recurGamma-as}) with 
$\mathbf{B}=\boldsymbol{\Gamma}_n$,  
$\mathbf{C}  =  \boldsymbol{\Gamma}_{n} - \frac{n}{n+1} \left(\mathbf{x}_{n+1} - \boldsymbol{\mu}_{n} \right)\left(\mathbf{x}_{n+1} - \boldsymbol{\mu}_{n} \right)^T$,
and  $\delta = -1/(n+1)$.
Define $\phi_{j,n} = \left( \mathbf{x}_{n+1}- \boldsymbol{\mu}_{n}\right)^T \mathbf{v}_{j,n}$ 
for $j=1,\ldots,d$. 
Assuming that the eigenelements $(\lambda_{j,n}, \mathbf{u}_{j,n})$ of  
$\boldsymbol{\Gamma}_n$ satisfy $\lambda_{1,n} > \ldots > \lambda_{d,n} $, 
we have 
\begin{align}
\lambda_{j,n+1} &\approx  \lambda_{j,n}  + \frac{1}{n+1} \left(  \frac{n}{(n+1)} \, \phi_{j,n}^2 
-  \lambda_{j,n} \right),
\label{perturb:valp}\\
\mathbf{u}_{j,n+1}  &\approx \mathbf{u}_{j,n}  + \frac{n}{(n+1)^2} \left( \sum_{i \neq j} \frac{\phi_{j,n} \phi_{i,n}}{\lambda_{j,n}- \lambda_{i,n}}\, \mathbf{u}_{i,n} \right).
\label{perturb:vecp}
\end{align}

% VERSION CENTERED ON MU_{N+1}
%We now apply Lemma \ref{lem:perturb} 
%to  (\ref{def:recurGamma-as}) with 
%$\mathbf{B}=\boldsymbol{\Gamma}_n$,  $\delta = -1/(n+1)$,
%and $\mathbf{C}  =  \boldsymbol{\Gamma}_{n} - \frac{n+1}{n} \left(\mathbf{x}_{n+1} - \boldsymbol{\mu}_{n+1} \right)\left(\mathbf{x}_{n+1} - \boldsymbol{\mu}_{n+1} \right)^T$. 
%We denote the eigenpairs of $\boldsymbol{\Gamma}_n$ by 
% $(\lambda_{j,n}, \mathbf{v}_{j,n}),\, j=1,\ldots,d,$ as before and 
%define $\phi_{j,n} = \left(  \mathbf{x}_{n+1}- \boldsymbol{\mu}_{n+1}\right)^T \mathbf{v}_{j,n}$. 
%Assuming that
%$\lambda_{1,n} > \ldots > \lambda_{d,n} $, 
%we have: % for $j=1, \ldots, d$:
%\begin{align}
%\lambda_{j,n+1} &\approx  \lambda_{j,n}  - \frac{1}{n+1} \left(  \lambda_{j,n} - \frac{n+1}{n} \, \phi_{j,n}^2 \right),
%\label{perturb:valp}\\
%\mathbf{v}_j^{(n+1)}  &\approx \mathbf{v}_j^{(n)}  + \frac{n}{(n+1)^2} \left( \sum_{i \neq j} \frac{\phi_{j,n} \phi_{i,n}}{\lambda_{j,n}- \lambda_{i,n}}\, \mathbf{v}_{i,n} \right).
%\label{perturb:vecp}
%\end{align}

  Similar approximations %based on perturbation arguments 
  can be found  in \cite{Lietal2000} and \cite{Hegdeetal2006}. 
Note that the assumption that $\boldsymbol{\Gamma}_n$ has distinct eigenvalues is not restrictive: indeed,  
the eigenanalysis of $\boldsymbol{\Gamma}_{n+1}$ can always be reduced to this case by deflation 
(see e.g.,   \cite{Bunchetal1978}).

\subsection*{Secular equations}

 \cite{GuEisenstat94} propose an exact and stable technique  for the case of a rank one perturbation  (see also the seminal work of \cite{Golub73}). For simplicity, suppose that $\mathbf{B}$ is a diagonal matrix with distinct eigenvalues $\lambda_1>  \ldots > \lambda_d$. The eigenvalues of the perturbed  matrix $\mathbf{B}(\delta) = \mathbf{B} + \delta  \mathbf{c}\mathbf{c}^T$, where $\mathbf{c} \in \mathbb{R}^d$ is taken to be an unit vector without loss of generality, can be computed exactly as the roots of the secular equation
\begin{align}
1 + \delta \sum_{j=1}^d \frac{c_j^2}{\lambda_j - \lambda} &= 0.
\end{align}
The eigenvectors of $\mathbf{B}(\delta)$ can then be computed exactly by applying 
 (\ref{perturb:vecp}) to the eigenvalues of $\mathbf{B}(\delta)$ and eigenvectors of $\mathbf{B}$.
%This rank one update framework corresponds exactly to the perturbation scheme (\ref{def:recurGamma1})  when the data arrive sequentially.

A major drawback of perturbation techniques for online PCA is that they require computing all eigenelements of the covariance matrix. Accordingly, for each new observation vector, $O(d^2)$ flops are needed to update the PCA and $O(d^2)$ memory is required to store the results. 
This computational burden and storage requirement are prohibitive for large $d$.

%%%%%%%%%%%%%%%%%%%%%%%%%%%%%%%%%%%%%%%%%%%%%%%%%%%%%%%%%%%%%%%

\section{Reduced rank incremental PCA}
\label{sec:incremental}

\cite{Aroraetal2012} suggest an incremental PCA (IPCA) approach based on the incremental SVD of  \cite{Brand02incrementalsingular}. In comparison to perturbation methods, a decisive advantage of IPCA is that it does not require computing all $d$ eigenelements if one is only interested in the $q<d$ largest eigenvalues. This massively  speeds up computations when $q $ is much smaller than $ d$. A limitation of the IPCA algorithm of \cite{Aroraetal2012} is that it only performs updates with respect to a single data vector. A similar procedure allowing for block updates can be found in \cite{LevyLindenbaum2000}.

Let $\boldsymbol{\Delta}_n = \mathbf{U}_n \mathbf{D}_n \mathbf{U}_n^T$   be a rank $q$ approximation to $\boldsymbol{\Gamma}_n$,  where the $q\times q$ diagonal matrix  $\mathbf{D}_n $ approximates the first $q$ eigenvalues of  $\boldsymbol{\Gamma}_n$ and the $d \times q$ matrix $\mathbf{U}_{n}$  approximates the corresponding eigenvectors of  $\boldsymbol{\Gamma}_n$. 
The columns of $\mathbf{U}_n$ are orthonormal  so that $\mathbf{U}_n^T\mathbf{U}_n = \mathbf{I}_q$.  
When a new observation $\mathbf{x}_{n+1} $ becomes available,  $\boldsymbol{\Delta}_n$ can be updated as follows. 
%First, the sample mean $\boldsymbol{\mu}_{n}$ is updated to $\boldsymbol{\mu}_{n+1}$ via \eqref{def:recurMean1}. 
The centered vector $\widetilde{\mathbf{x}}_{n+1} =   \mathbf{x}_{n+1} -  \boldsymbol{\mu}_{n} $ is decomposed as $\widetilde{\mathbf{x}}_{n+1}  = \mathbf{U}_n \mathbf{c}_{n+1} + \widetilde{\mathbf{x}}_{n+1}^{\perp}$, where $\mathbf{c}_{n+1} = \mathbf{U}_{n}^T \,\widetilde{\mathbf{x}}_{n+1} $ are the coordinates of  $\widetilde{\mathbf{x}}_{n+1} $ in the $q$-dimensional space spanned by $\mathbf{U}_n$ and $\widetilde{\mathbf{x}}_{n+1}^{\perp}$ is the projection of $\widetilde{\mathbf{x}}_{n+1}$ onto the orthogonal space of $\mathbf{U}_n$. 
In view of  \eqref{def:recurGamma1} the covariance $\boldsymbol{\Gamma}_{n+1}$ is approximated by $\boldsymbol{\Delta}_{n+1}= \frac{n}{n+1}\boldsymbol{\Delta}_n + \frac{n}{(n+1)^2} \widetilde{\mathbf{x}}_{n+1} \widetilde{\mathbf{x}}_{n+1}^T$ (note that $\boldsymbol{\Delta}_{n+1}$ is not  computed in the algorithm). 
The previous equation  rewrites as  
\begin{align}\label{def:Qmat}
%\left\{
%\begin{array}{ l  }
%\vspace*{2mm}
\boldsymbol{\Delta}_{n+1} & =
 \left[ \mathbf{U}_{n} \  \displaystyle\frac{\widetilde{\mathbf{x}}_{n+1}^{\perp} }{ \nrm{\widetilde{\mathbf{x}}_{n+1}^{\perp}}} \right]
\mathbf{Q}_{n+1} 
 \left[ \mathbf{U}_{n} \ \displaystyle \frac{\widetilde{\mathbf{x}}_{n+1}^{\perp} }{ \nrm{\widetilde{\mathbf{x}}_{n+1}^{\perp}}} \right]^T \\
 \noalign{\noindent with} %\nonumber
\mathbf{Q}_{n+1} & = \displaystyle  \frac{n}{(n+1)^2} \left(
\begin{array}{cc}
(n+1)\mathbf{D}_{n} +  \mathbf{c}_{n+1}\mathbf{c}_{n+1}^T& \nrm{\widetilde{\mathbf{x}}_{n+1}^{\perp}}\mathbf{c}_{n+1} \\
\nrm{\widetilde{\mathbf{x}}_{n+1}^{\perp}}\mathbf{c}_{n+1}^T  & \nrm{\widetilde{\mathbf{x}}_{n+1}^{\perp}}^2 \end{array}
\right).
\end{align}
%with 
%\begin{equation}
%\mathbf{Q}_{n+1}  =  \frac{n}{(n+1)^2} \left(
%\begin{array}{cc}
%(n+1)\mathbf{D}_{n} +  \mathbf{c}_{n+1}\mathbf{c}_{n+1}^T& \nrm{\widetilde{\mathbf{x}}_{n+1}^{\perp}}\mathbf{c}_{n+1} \\
%\nrm{\widetilde{\mathbf{x}}_{n+1}^{\perp}}\mathbf{c}_{n+1}^T  & \nrm{\widetilde{\mathbf{x}}_{n+1}^{\perp}}^2 
%\end{array}
%\right),
%\end{equation}

It then suffices to perform the EVD of the matrix  
$\mathbf{Q}_{n+1} $ of dimension $(q+1) \times (q+1)$. 
Writing $\mathbf{Q}_{n+1}  =  \mathbf{V}_{n+1} \mathbf{S}_{n+1}\mathbf{V}_{n+1}^T$ 
with $\mathbf{V}_{n+1}$ orthogonal and $\mathbf{S}_{n+1}$ diagonal, 
the EVD of %$(\mathbf{U}_{n+1} ,\mathbf{D}_{n+1})$ 
 $\boldsymbol{\Delta}_{n+1} $ simply expresses as $ \mathbf{U}_{n+1} \mathbf{D}_{n+1} \mathbf{U}_{n+1}^T$, where 
$\mathbf{D}_{n+1}   = \mathbf{S}_{n+1} $ and 
\begin{equation} 
\mathbf{U}_{n+1}  = \left[ \mathbf{U}_{n} \  \
\frac{\widetilde{\mathbf{x}}_{n+1}^{\perp} }{ \nrm{\widetilde{\mathbf{x}}_{n+1}^{\perp}}} \right] \mathbf{V}_{n+1}. 
\end{equation}
To keep the approximation $\boldsymbol{\Delta}_{n+1} $ of $\boldsymbol{\Gamma}_{n+1}$ at rank $q$, 
 the row and column of $\mathbf{D}_{n+1}$ containing the smallest eigenvalue are deleted and    
the associated eigenvector is deleted from $\mathbf{U}_{n+1}$.

\section{Stochastic approximation} 
\label{sec: stochastic}

\subsection{Stochastic gradient optimization}
\label{sec: stochastic gradient}

Stochastic gradient approaches adopt a rather different point of view based on the population version of the optimization problem  (\ref{def:riskPCA}). 
Stochastic gradient algorithms for online PCA  have been  proposed by 
\cite{Sanger1989}, \cite{Krasulina1970}, \cite{OjaKarhunen1985} and \cite{Oja1992}.
Parametric convergence rates for the first eigenvalue and eigenvector have been obtained recently by \cite{BalsubramaniDF2013}.
These algorithms are very fast and differ mostly in how they (approximately) orthonormalize eigenvectors after each iteration.

Let $\mathbf{X}$ be a random  vector taking values in $\R^d$ with mean  
%We assume that $\mathbb{E}\left( \| \mathbf{X}\| ^2\right) < \infty$ 
%%, where $\nrm{\cdot}$ is the Euclidean norm, 
%and write 
$\boldsymbol{\mu} = \mathbb{E}(\mathbf{X})$ and covariance $\boldsymbol{\Gamma} = \mathbb{E}\big[ \left(\mathbf{X}-\boldsymbol{\mu} \right)\left(\mathbf{X}-\boldsymbol{\mu} \right)^T \big]$.  
Consider the minimization of the compression loss 
\begin{align}\label{riskPCA:population}
R(\mathbf{P}_q) &= \EE{ \nrm{\left( \mathbf{X}  - \boldsymbol{\mu} \right) - \mathbf{P}_q \left( \mathbf{X}  - \boldsymbol{\mu}  \right)}^2} \\ 
& = \mathrm{tr}\left(  \boldsymbol{\Gamma} \right)- \mathrm{tr}\left(   \mathbf{P}_q  \boldsymbol{\Gamma} \right) ,  \nonumber
\end{align}
where $\mathbf{P}_q$ is a projection matrix onto some $q$-dimensional subspace of $\R^d$. 
Note that \eqref{riskPCA:population} is the probabilistic version of the empirical loss \eqref{def:riskPCA}. 

Let $\mathbf{U}$ be a $d\times q$ matrix whose columns form an orthonormal basis of the projection subspace, 
so that $\mathbf{P}_q = \mathbf{UU}^T$. 
The minimization of \eqref{riskPCA:population} is conveniently reformulated as the maximization  of 
\begin{equation}\label{compression criterion: trace}
\Psi (\mathbf{U}) =\mathrm{tr}\,( \boldsymbol{\Gamma}\mathbf{UU}^T )
\end{equation}
whose gradient  is 
\begin{equation}\label{psi gradient}
\nabla \Psi (\mathbf{U}) = 2\, \boldsymbol{\Gamma}  \mathbf{U}.
\end{equation}

Assuming for now that $\boldsymbol{\Gamma}$ is known and ignoring the orthonormality constraints on $\mathbf{U}$,  a gradient ascent algorithm would have updates of the form $\mathbf{U}_{n+1} = \mathbf{U}_{n} + \gamma_n \ \boldsymbol{\Gamma} \mathbf{U}_{n}$ with $\gamma_n$  a step size. 
Since $\boldsymbol{\Gamma}$ is in fact unknown, it is replaced by a random approximation whose expectation is proportional to $\boldsymbol{\Gamma}$.
%In stochastic gradient ascent, the gradient, which is here proportional to ,. It to the gradient.
Accordingly, for each new observation $\mathbf{x}_{n+1}$, the matrix $\mathbf{U}_n$ of orthonormal vectors 
%recursive approximation to the $j$th eigenvector, for $j=1, \ldots, q$, 
is updated as follows:
\begin{align}
\tilde{\mathbf{U}}_{n+1}&= \mathbf{U}_n + \gamma_{n} \left(\mathbf{x}_{n+1}- \boldsymbol{\mu}_{n+1} \right)\left(\mathbf{x}_{n+1}- \boldsymbol{\mu}_{n+1} \right)^{T} \mathbf{U}_n,
\label{def:stochPCA1} \\
\mathbf{U}_{n+1} &= \mbox{Orthonormalization} \big(\tilde{\mathbf{U}}_{n+1} \big) ,
\label{orthonormalization}
\end{align}
where %the sequence 
$(\gamma_n)$ satisfies the usual conditions %for convergence 
of Robbins-Monro algorithms: 
 $\sum_{n \geq 1} \gamma_n^2 < \infty$ and $\sum_{n \geq 1} \gamma_n = \infty$  \citep[e.g.,][]{Duf97}. The first condition ensures the almost sure convergence of the algorithm whereas the second guarantees convergence to the global maximizer of \eqref{compression criterion: trace}, namely, the  eigenvectors associated to the  $q$ largest eigenvalues of $\boldsymbol{\Gamma}$.

The update \eqref{def:stochPCA1} of the projection space has computational complexity $O(qd)$, 
which is much less than the complexity $O(d^2)$ of perturbation techniques if $q \ll d$. 
The orthonormalization \eqref{orthonormalization} can be realized for example with the Gram-Schmidt procedure. 
In this case, \cite{Oja1992} describes the combination \eqref{def:stochPCA1}-\eqref{orthonormalization} as the Stochastic Gradient Ascent (SGA) algorithm. 
Although the Gram-Schmidt procedure requires $O(q^2d)$ elementary operations,
%However, as noted in \cite{Aroraetal2012}, 
the space generated by the estimated eigenvectors remains the same even if orthonormalization is not performed at each step. As a result, if orthonormalization is performed every $q$ steps or so, the overall complexity of the SGA algorithm remains $O(q d)$ per iteration. 
Alternatively, computational speed can be increased at the expense of numerical accuracy 
by using a first order approximation of the Gram-Schmidt orthonormalization (i.e., neglecting the terms of order $O(\gamma_n^2)$).  
This enables the   approximate implementation 
\begin{equation}\label{def:sga}
\mathbf{u}_{j,n+1} = \mathbf{u}_{j,n} + \gamma_n \phi_{j,n} \left[ \left( \mathbf{x}_{n+1}- \boldsymbol{\mu}_{n+1}\right) - \phi_{j,n} \mathbf{u}_{j,n} - 2 \sum_{i=1}^{j-1} \phi_{i,n} \mathbf{u}_{i,n} \right] ,
\end{equation}
where $\phi_{j,n} = \left( \mathbf{x}_{n+1}- \boldsymbol{\mu}_{n+1}\right)^T \mathbf{u}_{j,n}$. 
This fast implementation of the SGA algorithm can be interpreted as a neural network \citep{Oja1992}. %\cite{OjaKarhunen1985} and

The SGA algorithm also allows consistent recursive estimation of 
 the eigenvalues of $\boldsymbol{\Gamma}$  \citep[see][and \eqref{perturb:valp}]{OjaKarhunen1985}: 
%. For $j=1, \ldots, q$,
\begin{equation}
\lambda_{j,n+1}  = \lambda_{j,n}  + \gamma_n \left( \phi_{j,n}^2- \lambda_{j,n} \right).
\end{equation}
 
\medskip 

Subspace Network Learning (SNL) is another stochastic gradient algorithm in which   
the orthonormalization \eqref{orthonormalization} consists in multiplying $\tilde{\mathbf{U}}_{n+1}$ 
by $ \big(\tilde{\mathbf{U}}_{n+1}^T\tilde{\mathbf{U}}_{n+1} \big)^{-1/2}$  \citep{Oja1983,Oja1992}. 
Using first order approximations, a fast approximate implementation of SNL is
\begin{align}
\mathbf{u}_{j,n+1} &= \mathbf{u}_{j,n} + \gamma_n \phi_{j,n} \left[  \left( \mathbf{x}_{n+1}- \boldsymbol{\mu}_{n+1}\right)  -  \sum_{i=1}^q \phi_{i,n} \mathbf{u}_{i,n} \right].
\label{def:snl}
\end{align}
The SNL algorithm is faster than SGA but unlike SGA, it only converges to the eigenvectors of $\boldsymbol{\Gamma}$ up to a rotation. In other words, SNL recovers the eigenspace generated by the first $q$ eigenvectors but not the eigenvectors themselves. 

\cite{Sanger1989} proposes a neural network approach called the Generalized Hebbian Algorithm (GHA):
 \begin{equation}\label{def:gha}
\mathbf{u}_{j,n+1} = \mathbf{u}_{j,n} + \gamma_n \phi_{j,n} \left[ \left( \mathbf{x}_{n+1}- \boldsymbol{\mu}_{n+1}\right) - \phi_{j,n} \mathbf{u}_{j,n} - \sum_{i=1}^{j-1} \phi_{i,n} \mathbf{u}_{i,n} \right] .
\end{equation}
The almost sure convergence of the estimator $\mathbf{u}_{j,n}$  to the corresponding eigenvector of $\boldsymbol{\Gamma}$ for $j=1,\ldots,q,$ is established in the same paper. By construction, the vectors $\mathbf{u}_{j,n}, \, j=1,\ldots,q,$ are mutually orthogonal. In practice however, loss of orthogonality may occur  due to roundoff errors. 
As noted in \cite{Oja1992}, GHA is very similar to the fast implementation \eqref{def:sga} of SGA, the only difference being that there is no coefficient 2 in the sum. 
Strictly speaking, however, GHA is not a stochastic gradient algorithm.
%Another way to perform online and simultaneous orthonormalization of the approximated eigenvectors is suggested in  and . 
%
%\subsection{Generalized hebbian algorithms}

\subsection{Choosing the learning rate}

The choice of the learning rate sequence $(\gamma_n)_{n \geq 1}$  in the previous stochastic algorithms has great practical importance, yet is rarely discussed  in the literature. A usual choice is $\gamma_n = c/n$ for some well chosen constant $c$. However, if  $c$ is too small, the algorithm may get stuck far from the optimum whereas if $c$ is too large, it may have large oscillations. There are no universally good values for  $c$ because a sensible choice should depend on  the magnitude of the data vectors, the distance between the starting point of the algorithm and the global solution, the gaps between successive eigenvalues of $\boldsymbol{\Gamma}$,  etc.. In practice, $c$ is often selected by trial and error \citep[e.g.,][]{Oja1983}.

A more prudent strategy consists in using learning rates of the form $\gamma_n = c n^{-\alpha}$ with $\alpha \in (1/2, 1)$.  In this way,  $\gamma_n$ tends to zero less rapidly so that the algorithm is allowed to oscillate and has less chances to get stuck at a wrong position.  This is particularly important if the starting point of the algorithm is far from the solution.

%Better performances can be obtained if $c$ depends on the rank $j$ of the eigenvalue ($j=1,\ldots,q$). 
%%, so that we have a sequence of steps controlled by $c_1, \ldots, c_q$. 
%More precisely, when $j$ increases, larger values of $c=c_j$ tend to improve the performances of the algorithm. As noted in \cite{BalsubramaniDF2013} the tuning parameter $c_1$ should  satisfy  $c_1 \geq  0.5/(\lambda_1-\lambda_2)$ to ensure convergence of the SGA algorithm towards the largest eigenvector of $\boldsymbol{\Gamma}$ 
%at the optimal rate $O(1/n)$. One may conjecture that for $j>1$, the condition $c_j \ge  0.5/ \min ( \lambda_{j-1}-\lambda_j, \lambda_{j}-\lambda_{j+1}) $ should be satisfied to guarantee convergence at the optimal rate. 
%
%Averaging techniques may be interesting in that case to improve significantly the quality of the estimates (see \cite{PolyakJud92}), in particular if one is only interested in estimating the first eigenfunction (see the simulation study).

Data-driven methods have been proposed in the literature to select the learning rate. For example, a non-zero-approaching adaptive learning rate of the form $\gamma_n = c/ \mathbf{u}_n^T \boldsymbol{\Gamma}_n \mathbf{u}_n$ is studied in  \cite{Lv2006}, where $c$ is a constant to be chosen in $(0,0.8).$ 
In our simulations however, this technique produced poor performances not reported here.

\subsection{Candid covariance-free incremental PCA}

\cite{WZH2003} propose a method called candid covariance-free incremental PCA (CCIPCA) that resembles SGA and GHA. This method however does not aim to optimize an objective function or train a neural network. Following \cite{WZH2003}, we first present the algorithm in the case where $\boldsymbol{\mu} = \mathbf{0}$  and then consider the general case. 
Let  $\mathbf{u}$ be an eigenvector of $\boldsymbol{\Gamma}$ with unit norm and let $\lambda$ be the associated eigenvalue. Assume that estimates $\mathbf{v}_0,\cdots, \mathbf{v}_{n-1} $ of $\mathbf{v}=\lambda \mathbf{u}$ have been constructed in previous steps. 
The  idea of CCIPCA is to substitute $ \mathbf{x}_i  \mathbf{x}_i^T$ to $\boldsymbol{\Gamma}$ and $\mathbf{v}_{i-1} / \|\mathbf{v}_{i-1} \| $ to $\mathbf{u}$ in the eigenequation $\boldsymbol{\Gamma} \mathbf{u} = \lambda \mathbf{u}$
  for $i=1,\ldots, n,$ and to average the results: 
\begin{equation}\label{ccipca-iter}
 \mathbf{v}_{n} = \frac{1}{n} \sum_{i=1}^{n}  \mathbf{x}_i \mathbf{x}_i^T \, \frac{\mathbf{v}_{i-1}}{\|  \mathbf{v}_{i-1} \|} \,  .
\end{equation}
The normalized eigenvector $\mathbf{u}$ and eigenvalue $\lambda$ are estimated by $\mathbf{u}_{n} = \mathbf{v}_{n}/\| \mathbf{v}_{n}\|$ and $\| \mathbf{v}_{n}\|$, respectively. A proof of the almost sure convergence of $\mathbf{v}_{n}$ to $\mathbf{v}$ can be found in \cite{ZhangWeng2001}. 

As can be seen in \eqref{ccipca-iter}, CCIPCA produces a sequence of stochastic approximations to the eigenvectors  of $\boldsymbol{\Gamma} $ and then averages them. This is a major difference compared to the previous stochastic approximation algorithms that directly target the population covariance matrix. Because it is based on averaging, CCIPCA does not require specifying tuning parameters. This is a major advantage over SGA and GHA.  

From a computational standpoint, \eqref{ccipca-iter} is conveniently written in recursive form as 
\begin{equation}\label{def:CCIPCA}
\mathbf{v}_{n+1}  =  \frac{n}{n+1}\,\mathbf{v}_{n} + 
 \frac{1}{n+1} \, \mathbf{x}_{n+1}  \mathbf{x}_{n+1}^T \, \frac{\mathbf{v}_{n}}{\left\| \mathbf{v}_{n}\right\|}  \,. 
\end{equation}
The algorithm can be initialized by $\mathbf{v}_0 =  \mathbf{x}_{1} $. 
In the general case where $ \boldsymbol{\mu} $ is unknown, it should be estimated via \eqref{def:recurMean1} and $\mathbf{x}_{n+1}$ should be centered on $ \boldsymbol{\mu}_{n+1}$ in \eqref{def:CCIPCA} for $n\ge 1$  (note that $\mathbf{x}_1-\boldsymbol{\mu}_1=\mathbf{0}$). A suitable initialization is $\mathbf{v}_0=\mathbf{v}_1= \mathbf{x}_{1} - \boldsymbol{\mu}_2$. 

%Alternatively, this algorithm 
%\begin{equation}
%\mathbf{v}_{n+1}  = \mathbf{v}_{n} + \frac{1}{n+1}  \left( \mathbf{x}_{n+1}  \mathbf{x}_{n+1}^T 
%\frac{\mathbf{v}_{n}}{\left\| \mathbf{v}_{n}\right\|} - \mathbf{v}_{n} \right)
%\label{def:CCIPCA}
%\end{equation}

When estimating more than one eigenvector, say $\mathbf{v}_1,\ldots, \mathbf{v}_q$, the same deflation method as  in GHA is applied to enforce orthogonality of the estimates: to compute $ \mathbf{v}_{j+1,n}$,   the input vector $\mathbf{x}_{n+1}$ is replaced by $\mathbf{x}_{n+1} - \sum_{k=1}^j \big( \mathbf{x}_{n+1}^T \mathbf{u}_{k,n}\big) \mathbf{u}_{k,n}  $
 in \eqref{def:CCIPCA}. This saves much computation time compared to Gram-Schmidt orthonormalization but may cause loss of orthogonality due to roundoff errors.

To handle data generated by nonstationary processes, a parameter $\ell \ge 0$ called amnesic factor 
can be introduced in \eqref{def:CCIPCA}: 
\begin{align}
\mathbf{v}_{n+1} & = \frac{n-\ell }{n+1}\, \mathbf{v}_{n} + \frac{1+\ell}{n+1} \, \mathbf{x}_{n+1} \mathbf{x}_{n+1} ^T \,\frac{\mathbf{v}_{n}}{\left\| \mathbf{v}_{n}\right\|} \,  .
\label{def:CCIPCAell}
\end{align}
This parameter controls the weight given to earlier observations. According to \cite{WZH2003}, $\ell$ should typically range between $2$ and $4$, with larger values of $\ell$ giving more weight to recent observations. For $\ell=0$,  \eqref{def:CCIPCAell} reduces to the stationary case \eqref{def:CCIPCA}.

%%%%%%%%%%%%%%%%%%%%%%%%%%%%%%%%%%%%%%%%%%%%%%%%%%%%%%%%%%%%%%%

\section{Extensions}
\label{sec: extensions}

\subsection{Nonstationary processes}

The perturbation methods of Section \ref{sec:perturbation} and IPCA algorithm of Section \ref{sec:incremental} 
have been presented under the implicit assumption of a stationary data-generating process. 
However, these methods can easily handle nonstationary processes 
by generalizing the sample mean and sample covariance as follows:  
\begin{align} %  \label{def:recurmu nonstationary} 
\boldsymbol{\mu}_{n+1} & =\left(1-f\right) \boldsymbol{\mu}_{n} +  f\, \mathbf{x}_{n+1}, \\
\boldsymbol{\Gamma}_{n+1} & =\left(1-f\right) \boldsymbol{\Gamma}_{n} +  f\left(1-f\right) \left(\mathbf{x}_{n+1} - \boldsymbol{\mu}_{n} \right)\left(\mathbf{x}_{n+1} - \boldsymbol{\mu}_{n} \right)^T  ,
\end{align}
where $0<f < 1$ is a "forgetting factor" that determines the weight of a new observation in the mean and covariance updates. In the stationary case, equations \eqref{def:recurMean1}-\eqref{def:recurGamma1} are recovered by 
setting $f=1/(n+1)$. More generally, larger values of $f$ give more weight on new observations. 

The stochastic algorithms of Section \ref{sec: stochastic gradient} naturally accommodate nonstationary processes 
through the learning rate $\gamma_n$.

\subsection{Missing data}
\label{sec: missing data}

Standard imputation methods (mean, regression, hot-deck, maximum likelihood, multiple imputation, etc.) can be used to handle missing data in the context of online PCA. 
See e.g., \cite{Josseetal2011} for a description of missing data imputation in offline PCA. 

Hereafter we describe the approach of  \cite{Brand02incrementalsingular} that imputes missing values in the observation vector $\mathbf{x}_{n+1}$ by empirical best linear unbiased prediction (EBLUP). 
The key idea  is to consider $\mathbf{x}_{n+1}$ as a realization of the multivariate normal distribution $ N_d (\boldsymbol{\mu}_{n}, \mathbf{U}_n \mathbf{D}_n \mathbf{U}_n^T)$. In other words, the population mean vector $\boldsymbol{\mu}$ and covariance matrix $\boldsymbol{\Gamma}$ are approximated using the current sample mean vector and PCA. 

Partition  $\mathbf{x}_{n+1}$ into two subvectors $\mathbf{x}_{n+1}^o$ (observed values) and $\mathbf{x}_{n+1}^m$ (missing values) of respective sizes $d-m_n$ and $m_n$, where $0 < m_n < d$ is the number of missing values. 
Similarly, partition $\boldsymbol{\mu}_{n}$ in two subvectors $\boldsymbol{\mu}_{n}^o$ and $\boldsymbol{\mu}_{n}^m$ 
 whose entries correspond to $\mathbf{x}_{n+1}^o$ and $\mathbf{x}_{n+1}^m$. 
Also partition $\mathbf{U}_n$ in two submatrices $\mathbf{U}_{n}^o$ and $\mathbf{U}_{n}^m$ of respective dimensions $(d-m_n)\times q$ and $m_n \times q$ whose rows correspond to $\mathbf{x}_{n+1}^o$ and $\mathbf{x}_{n+1}^m$. Let 
$\mathbf{D}_n^{1/2}$ be the diagonal matrix containing the square roots of the diagonal elements of $\mathbf{D}_n$.
 By the properties of conditional expectations for multivariate normal distributions,  
the empirical best linear unbiased predictor of $\mathbf{x}_{n+1}^m$ is
\begin{equation}
\widehat{\mathbf{x}}_{n+1}^m 
= \mathbb{E} \left( \mathbf{x}_{n+1}^m  \big|  \mathbf{x}_{n+1}^o  , \boldsymbol{\mu}_n , \mathbf{U}_n,\mathbf{D}_n\right) =  \boldsymbol{\mu}_{n}^m + 
  \big( \mathbf{U}_n^m \mathbf{D}_n^{1/2} \big) \big(  \mathbf{U}_n^o \mathbf{D}_n^{1/2} \big)^+ 
  \left( \mathbf{x}_{n+1}^o - \boldsymbol{\mu}_{n}^o \right).
\label{def:imputincRpca}
\end{equation}

%%%%%%%%%%%%%%%%%%%%%%%%%%%%%%%%%

%%% Projecting functional data
\subsection{Functional data}
\label{fpca}
 
In many applications, the data are functions of a continuous argument (e.g., time, space, or frequency) observed on a dense grid of points $a \leq t_1 < \ldots < t_d \leq b$. 
The corresponding observation vectors $\mathbf{x}_i = (x_i(t_1), \ldots, x_i(t_d) )\in \mathbb{R}^d, \, i=1,\ldots, n,$ are often high-dimensional. 
Rather than  carrying out PCA directly on the $ \mathbf{x}_i, $ it is advantageous to consider the functional version of this problem (FPCA).
FPCA consists in finding the eigenvalues and eigenfunctions of the linear operator $T_n: \phi \in L^2([a,b]) \mapsto \big(  t\mapsto  \int_a^b \Gamma_n (s, t ) \phi (s)ds\big) $ associated to the  empirical covariance function
\[ \Gamma_n (s,t) = \frac{1}{n} \sum_{i=1}^n \left(x_i(s) - \mu_n(s)\right)\left(x_i(t)-\mu_n(t)\right), \] 
where $\mu_n(t) = n^{-1} \sum_{i=1}^n x_i(t)$ the empirical mean function. 
By accounting for the structure of the data (e.g., time ordering) and the  smoothness of the eigenfunctions, 
FPCA can both reduce data dimension, hence computation time, and increase statistical accuracy.

An efficient way to implement FPCA is to first approximate the functions $x_i$ in a low-dimensional  space: 
\begin{equation}\label{basis approx}
x_i(t) \approx  \sum_{j=1}^p \beta_{ij} B_j(t),
\end{equation}
where $B_1, \ldots, B_p$ are smooth basis functions and $p \ll d$. 
%Calling $\tilde{x}_i$ the basis function approximation of $x_i$ in, 
The FPCA of the approximate $x_i$ in \eqref{basis approx} now reduces to the PCA of the basis coefficients $\boldsymbol{\beta}_i =(\beta_{i1}, \ldots, \beta_{ip}) \in \R^p$ in the metric $\mathbf{M} = ( \langle B_j, B_k \rangle ) \in \R^{p \times p}$, which is the Gram matrix of the basis functions. More precisely, writing $\overline{\boldsymbol{\beta}}_n = n^{-1} \sum_{i=1}^n \boldsymbol{\beta}_i$, 
% spectral decomposition of the covariance operator $\Gamma_n $
it suffices to diagonalize the matrix
\begin{align*} 
\boldsymbol{\Delta}_n \mathbf{M} &= \frac{1}{n} \sum_{i=1}^n \left( \boldsymbol{\beta}_i - \overline{\boldsymbol{\beta}}_n\right)\left( \boldsymbol{\beta}_i - \overline{\boldsymbol{\beta}}_n\right)^T \mathbf{M}
\end{align*}
with eigenvectors $\boldsymbol{\phi}_{j,n} \in \mathbb{R}^p, \, j=1, \ldots,p,$ satisfying the orthonormality constraints 
\begin{align}\label{M-ortho}
\boldsymbol{\phi}_{j,n}^T \mathbf{M} \boldsymbol{\phi}_{\ell,n} &= \delta_{j\ell}.
\end{align} 
Further details can be found in \cite{Ramsay_Silverman_Livre}. %, Chapter 8). 

In turn, the eigenvectors $\boldsymbol{\phi}_{j,n}$ can be found through the eigenanalysis of the symmetric matrix $\mathbf{M}^{1/2} \boldsymbol{\Delta}_n \mathbf{M}^{1/2}$. % \hnote{Decomposition de Choleski  de M ?} 
Indeed, if $\tilde{\boldsymbol{\phi}}_{j,n}$ is a (unit norm) eigenvector of  $\mathbf{M}^{1/2} \boldsymbol{\Delta}_n \mathbf{M}^{1/2}  $ associated to the eigenvalue $\lambda_{j,n}$, then $\boldsymbol{\phi}_j = \mathbf{M}^{-1/2}\tilde{\boldsymbol{\phi}}_j$ is an eigenvector of $\boldsymbol{\Delta}_n \mathbf{M}$ for the same eigenvalue and the $\boldsymbol{\phi}_j, \, j=1,\ldots, p,$ satisfy \eqref{M-ortho}.  It is possible to obtain a reduced-rank FPCA by computing only $q<p$ eigenvectors, but this is not as crucial as for standard PCA because the data dimension has already been reduced from $d$ to $p\ll d$.   

To extend FPCA to the online setup, the following steps are required: 

\begin{enumerate}
\item Given a new observation $\mathbf{x}_{n+1} \in \mathbb{R}^d$, compute %its basis coefficients 
$\boldsymbol{\beta}_{n+1} = \mathbf{A}\mathbf{x}_{n+1} \in \mathbb{R}^p$, where $\mathbf{A}$ is a suitable matrix. Typically, $\mathbf{A}=\left( \mathbf{B}^T \mathbf{B}+\alpha \mathbf{P}\right)^{-1}\mathbf{B}^T$ with $\mathbf{B} = (B_j(t_k))$ of dimension $d\times p$, $\mathbf{P}$ a penalty matrix, and $\alpha \ge 0$ a smoothing parameter. 

%Given the matrix $\mathbf{A}$, this step requires $O(pd)$ flops.
\item %Apply the $\mathbf{M}^{1/2}$ 
Update the sample mean  $\overline{\boldsymbol{\beta}}_{n}$ via \eqref{def:recurMean1} 
 and 
compute $\widetilde{\boldsymbol{\beta}}_{n+1} = \mathbf{M}^{1/2}\left( \boldsymbol{\beta}_{n+1} - \overline{\boldsymbol{\beta}}_{n+1} \right)$. 
%This steps requires $O(p^2)$ operations.
\item Apply an online PCA algorithm to update the eigenelements $(\lambda_{j,n}, \tilde{\boldsymbol{\phi}}_{j,n} )$ with respect to $\widetilde{\boldsymbol{\beta}}_{n+1}$. Note that the PCA takes place in $\mathbb{R}^p$ 
and not $\mathbb{R}^d$. 
\item Compute the eigenvectors $\boldsymbol{\phi}_{j,n+1} = \mathbf{M}^{-1/2} \tilde{\boldsymbol{\phi}}_{j,n+1}$. If needed, compute 
%expand the  $\boldsymbol{\phi}_{j,n+1} $ in the function basis %$(B_1, \ldots, B_p)$ 
%to obtain 
the discretized eigenfunctions 
$\mathbf{B}\boldsymbol{\phi}_{j,n+1} \in \mathbb{R}^d$.
\end{enumerate}

Steps 1 and 2 can be gathered for computational efficiency, that is, the matrix product $\mathbf{M}^{1/2}\mathbf{A}$ 
can be computed once for all and be directly applied to new data vectors.  This combined step requires $O(pd)$ flops. The time complexity of step 3 depends on the online PCA algorithm used; it is for example $O(q^2p)$ with IPCA and $O(qp)$ with GHA. If $q=O(p)$, one can also explicitly compute and update the covariance matrix $\boldsymbol{\Delta}_n$ with \eqref{def:recurMean1}, and perform its batch PCA for each $n$. The cost per iteration of this approach is the same as online PCA, namely $O(p^3)$. Step 4 produces the eigenvectors  $\boldsymbol{\phi}_{j,n+1} \in \mathbb{R}^p$ in $O(pq)$ flops and requires $O(qd)$ additional flops to compute eigenfunctions. Using for example IPCA in step 3, the total cost per iteration of FPCA is $O(pd)$, which makes it very competitive with standard (online) PCA. 
In addition, if  $p \ll d$ and the eigenfunctions of the covariance operator $T_n$ are smooth, 
FPCA can greatly improve the accuracy of estimates thanks to the regularized projection \eqref{basis approx} onto smooth basis functions. This fact is confirmed in the simulation study.

A standard choice for the penalty is $\mathbf{P} = (\langle B_j'', B_k'' \rangle)$, 
which penalizes curvature in the basis function approximation \eqref{basis approx}. 
The parameter $\alpha$ can be selected manually using pilot data. 
Alternatively, an effective automated selection procedure is to randomly split pilot data in two subsets 
and select the value $\alpha$ for which the FPCA of one subset (i.e., the projection matrix $\mathbf{P}_q(\alpha)$)  minimizes the loss function \eqref{def:riskPCA} for the other subset.

%%%%%%%%%%%%%%%%%%%%%%%%%%%%%%%%%%%%%%%%%%%%%%%%%%%%%%%%%%%%%%%

%%%% Comparison of the different approaches
\section{Comparison of online PCA algorithms}
\label{sec: simulations}

%\subsection{Number of elementary operations and required memory}
\subsection{Time and space complexity}

\begin{table}[h!]
\caption{Computational cost and memory usage of online PCA per iteration
%For the SGA algorithm,  ``ortho." and ``nn." refer to the exact  (with Gram-Schmidt orthonormalization) and neural network implementation, respectively. 
}
\begin{center}
\begin{tabular}{lcc} \hline
Method & Required Memory & Computation time \\ \hline 
Batch (EVD) &  $O(nd) $ &  $O(nd \min(n,d))$ \\  
Batch (SVD) & $O(nd)$ & $O(ndq)$ \\ 
SGA (ortho.) & $O(qd)$ &  $O(  q^2d)$  \\ 
SGA (nn) & $O(qd)$ &  $O(  qd)$  \\ 
GHA &  $O(qd)$ &  $O(  qd)$ \\
CCIPCA & $O(qd)$ &  $O(  qd)$ \\
Perturbation Approximation & $O(d^2)$ & $O(d^2)$ \\ 
%Rank $q$  Perturbation Approximation & $O(d^2)$ & $O(qd)$ ? \\ \hline 
Incremental PCA & $O(qd)$ & $O( q^2d)$  \\ \hline 
\end{tabular}
\end{center}
\label{table:complexity}
\end{table}%
Table \ref{table:complexity} compares the time and space complexity of the batch and online PCA algorithms under study.  
The usual batch PCA (EVD) does not scale with the data as it requires  $O(nd \min(n ,d))$ time. In comparison, truncated SVD has a computational cost that grows linearly with the data and hence can  be used with fairly large datasets. 
When $n$ is small, batch PCA (EVD or SVD) can provide reasonable starting points to online algorithms. 
If the dimension $d$ is large, perturbation methods  are very slow and require a large amount of memory. 
At the opposite end of the spectrum, the stochastic algorithms SGA and SNL (with neural network implementation - ``nn." in the table), GHA, and CCIPCA provide very fast PCA updates ($O(qd)$) with minimal memory requirement $O(qd)$ (this is the space needed to store the $q$ eigenvectors and eigenvalues). 
If $q$ is relatively small compared to $n$ and $d$, SGA and SNL (with exact orthonormalization - ``ortho." in the table) and IPCA offer efficient PCA updates ($O(q^2d)$ time complexity) albeit slightly slower than the previously mentioned stochastic algorithms.

\subsection{Simulation study}

\subsubsection{Setup}

A simulation study was conducted to compare the numerical performances of the online PCA algorithms.  
The data-generating model used for the simulation was a Gaussian random vector $\mathbf{X} $ in $\mathbb{R}^d$ with zero mean and covariance matrix $\boldsymbol{\Gamma} = \left( \min ( k , l)/d \right)_{ 1\le k,l \le d}$. 
This random vector can be interpreted as a Brownian motion observed at $d$ equidistant time points in $[0,1]$. 
For $d$ large enough, the eigenvalues of the scaled covariance $\boldsymbol{\Gamma}/d$  decrease rapidly   to zero ($\lambda_j \sim (j-0.5)^{-2}$) so that most of the variability of $\mathbf{X}$ is concentrated in a low-dimensional subspace of $\mathbb{R}^d$  \citep[e.g.,][]{AshGardner1975}.
In each simulation a number $n$ of independent realizations of $\mathbf{X}$ was  
generated  with  $n \in \{ 500, 1000\}$ and $d \in \{ 10,100,1000\} $. The online PCA algorithms were initialized by the batch PCA of the first $n_0=250$ observations and then run on the remaining $(n-n_0)$ observations. The number $q$ of estimated eigenvectors varied in $\{2,5,10,100\}$. 

%The estimation error for the $j$th eigenvalue $\lambda_j>0$ is evaluated by considering the relative absolute error
%\begin{align}
%R_j(\widehat{\lambda}_j) &=  \frac{ |\widehat{\lambda}_j - \lambda_j |}{ \lambda_j}.
%\label{def:errvalp}
%\end{align}

To evaluate the statistical accuracy of the algorithms, we considered the relative error in the estimation of the eigenspace  associated to the $q$ largest eigenvalues of $\boldsymbol{\Gamma}$. Let $\mathbf{P}_q = \mathbf{UU}^T$ be the  orthogonal projector on this eigenspace. 
Given a matrix $\widehat{\mathbf{U}}$ of estimated eigenvectors such that $\widehat{\mathbf{U}}^T\widehat{\mathbf{U}} = \mathbf{I}_q$, we consider the orthogonal projector  $\widehat{\mathbf{P}}_q = \widehat{\mathbf{U}}\widehat{\mathbf{U}}^T $ and measure the eigenspace estimation error by 
\begin{align}\label{def:errvecp} 
L(\widehat{\mathbf{P}}_q) &=  \big\| \widehat{\mathbf{P}}_q - \mathbf{P}_q  \big\|_F^2     
\big/ \big\| \mathbf{P}_q  \big\|_F^2 \\ 
 &= 2 \big( 1 - \mathrm{tr} \big[ \widehat{\mathbf{P}}_q \mathbf{P}_q  \big]  /q \big)\, , \nonumber
\end{align}
 where $\| \cdot \|_F $ denotes the Frobenius norm. 
In unreported simulations we also used the cosine between the top eigenvector of $\boldsymbol{\Gamma}$ and its estimate as a performance criterion and obtained qualitatively similar results to those presented here.

%We also consider the normalized version of the compression loss \eqref{riskPCA:population}:
%\begin{align*}\label{def:errcom}
%R(\widehat{\mathbf{P}}_q) &= 
%%\frac{
%\mathbb{E}_{\mathbf{X}} \Big[ \big\| \mathbf{X} -  \widehat{\mathbf{P}}_q \mathbf{X} \big\|^2  \Big] \Big/ \,
%\mathbb{E}_{\mathbf{X}} \big[ \big\| \mathbf{X}  \big\|^2 \big]
%%\mathrm{tr}\left(\boldsymbol{\Gamma}\right)
%\\
%%}
%%{ \\
%%{ \mathrm{tr}\left(\boldsymbol{\Gamma}\right)} \\
%& = % \frac{
%1 - \mathrm{tr}\big(\widehat{\mathbf{P}}_q \boldsymbol{\Gamma}\big)\big/ \, \mathrm{tr}\left(\boldsymbol{\Gamma}\right) .
%%}{\mathrm{tr}\left(\boldsymbol{\Gamma}\right) }. 
%\nonumber 
%\end{align*}
%The minimum of  $R$ is attained for $\widehat{\mathbf{P}}_q = \mathbf{P}_q$, 
%in which case $R(\mathbf{P}_q) = \sum_{j=q+1}^d \lambda_j / \sum_{j=1}^d \lambda_j$ is the percentage of unexplained variance after projecting the (centered) data onto the space generated by the $q$ eigenvectors of $\boldsymbol{\Gamma}$ associated to the $q$ largest eigenvalues.

\subsubsection{Computation time}

Computation times (in milliseconds, for one iteration) evaluated on a personal computer  (1,3 GHz Intel Core i5, with 8GB of RAM) are presented in Table~\ref{tab:comptempscalcul}. The results are globally coherent with Table~\ref{table:complexity}.
When the dimension $d$ of the data is small, all considered methods have comparable computation times 
at the exception of the secular equation approach which is at least ten times slower than the others.  
% However, exact perturbation approaches based on the resolution of a secular equation are at least algorithms (which give approximate solutions). 
As $d$ increases, the perturbation techniques, which compute all  eigenelements, get slower and slower compared to the other algorithms. For example, if we are interested in the first $q=5$ eigenvectors of a $1000 \times 1000$ covariance matrix, the CCIPCA and GHA algorithms  are more than 500 times faster than the perturbation approaches. 
The effect of the orthonormalization step on the computation time becomes much larger for high dimension $d$ and a relatively large number of computed eigenvectors $q$. When $d=1000$ and $q=100$, the GHA and CCIPCA algorithms that perform approximate orthonormalization are about seven times faster than IPCA and SGA that perform exact Gram-Schmidt orthonormalization.

\begin{table}[h]
\caption{Computation time (in milliseconds per iteration) of the online PCA algorithms}
\begin{center}
\begin{tabular}{lrrcrrcrrr}
\hline\hline
\multicolumn{1}{l}{\ }&\multicolumn{2}{c}{\ $d=10$}&\multicolumn{1}{c}{\ }&\multicolumn{2}{c}{\ $d=100$}&\multicolumn{1}{c}{\ }&\multicolumn{3}{c}{\ $d=1000$}\tabularnewline
\cline{2-3} \cline{5-6} \cline{8-10}
\multicolumn{1}{l}{}&\multicolumn{1}{c}{$q=2$}&\multicolumn{1}{c}{$q=5$}&\multicolumn{1}{c}{}&\multicolumn{1}{c}{$q=5$}&\multicolumn{1}{c}{$q=20$}&\multicolumn{1}{c}{}&\multicolumn{1}{c}{$q=5$}&\multicolumn{1}{c}{$q=20$}&\multicolumn{1}{c}{$q=100$}\tabularnewline
\hline
SGA (ortho.)&$0.12$&$0.09$&&$ 0.10$&$ 0.13$&&$   0.17$&$   0.77$&$  15.26$\tabularnewline
SGA (nn)&$0.09$&$0.09$&&$ 0.16$&$ 0.10$&&$   0.12$&$   0.37$&$   2.20$\tabularnewline
SNL (ortho.)&$0.14$&$0.16$&&$ 0.19$&$ 0.51$&&$   0.25$&$   1.46$&$  26.77$\tabularnewline
SNL (nn)&$0.08$&$0.12$&&$ 0.10$&$ 0.09$&&$   0.10$&$   0.31$&$   2.40$\tabularnewline
GHA&$0.05$&$0.06$&&$ 0.08$&$ 0.07$&&$   0.09$&$   0.36$&$   2.30$\tabularnewline
CCIPCA &$0.06$&$0.06$&&$ 0.07$&$ 0.18$&&$   0.13$&$   0.44$&$   2.04$\tabularnewline
IPCA&$0.08$&$0.17$&&$ 0.11$&$ 0.30$&&$   0.17$&$   1.00$&$  15.80$\tabularnewline
Perturbation (approx.)&$0.09$&$0.08$&&$ 1.81$&$ 1.50$&&$1221.35$&$1167.27$&$1197.58$\tabularnewline
Perturbation (secular)&$1.49$&$1.36$&&$17.26$&$18.81$&&$1515.46$&$1532.43$&$1481.94$\tabularnewline
\hline
\end{tabular}\end{center}
\label{tab:comptempscalcul}
\end{table}

\subsubsection{Statistical accuracy}

Table~\ref{tab:errvecp} shows the eigenspace estimation error $L$ %\eqref{def:errvecp} 
(averaged over 100 to 500 replications) of the online PCA algorithms for the first $q=5$ eigenvectors of $\boldsymbol{\Gamma}$ and different values of $n$ and $d$. To increase statistical accuracy, each algorithm actually computed $2q$ eigenvectors; only the first $q$ eigenvectors were kept for estimation in the end.  It is in  general advisable to compute more eigenvectors than required to maintain good accuracy for all target eigenvectors. 

The approximate perturbation approach produces estimation errors that are much greater at the end (that is, after all $n$ observations have been processed) than at the beginning (initialization by  batch PCA of $n_0=250$ observations). Therefore, this approach should not be used in practice. In contrast, the exact perturbation algorithm based on secular equation performs as well as batch PCA. 

The convergence of stochastic algorithms largely depends on the sequence of learning rates $\gamma_n$. 
Following the literature, we considered learning rates of the form $\gamma_n = c/ n^{\alpha}$, with $c>0$ and $\alpha \in (0.5,1]$. Larger values of $\alpha$ can be expected to produce better convergence rates but also increase the risk to get stuck close to the starting point of the algorithm. 
This can lead to poor results if the starting point is  far from the true eigenvectors of $\boldsymbol{\Gamma}$.
In our setup we use the values $\alpha \in \{1, 2/3\}$ and obtain the constants $b$ by minimizing 
the eigenspace estimation error $L$ over the grid $\{.01,.1,1,10,100\}$ (see Table~\ref{Table:learnrates}). 
Interestingly, smaller values of $c$ are chosen when the dimension $d$ increases and when $\alpha$ is decreases. 
The sample size $n$ does not seem to strongly impact the optimal value of $c$.
With our calibrated choice of the constant $c$ in the learning rates, the SGA and GHA algorithms display virtually identical performances. In addition, there is no great difference for the estimation error between $\alpha=1$ and $\alpha=2/3$. Given its high speed of computation, CCIPCA performs surprisingly well in all situations. IPCA is even more accurate and, although it is an approximate technique, it performs nearly as well as exact methods in this simulation study.

\begin{table}[htdp]
\caption{Best constant $c$ for the learning rate $\gamma_n=c/n^\alpha $ of the  SGA and GHA  algorithms. 
The best constants are identical for the two methods}
\begin{center}
\begin{tabular}{l c c c c c c c}
\hline
\hline
& \multicolumn{3}{c}{  $\alpha=1$} & \multicolumn{1}{c}{\ } & \multicolumn{3}{c}{  $\alpha=2/3$} \\ 
\cline{2-4} \cline{6-8} 
 & $d=10$ & $d=100$ & $d=1000$ && $d=10$ & $d=100$ & $d=1000$ \\
\hline
$n=500 $ & 10 &1 &0.1 && 1 &0.1 &0.01  \\
$n=1000 $ & 10 & 1 &0.1  && 1 & 0.1 & 0.01 \\
\hline
\end{tabular}
\end{center}
\label{Table:learnrates}
\end{table}

\begin{figure}[h]
   \begin{center}
  \includegraphics[width=14cm]{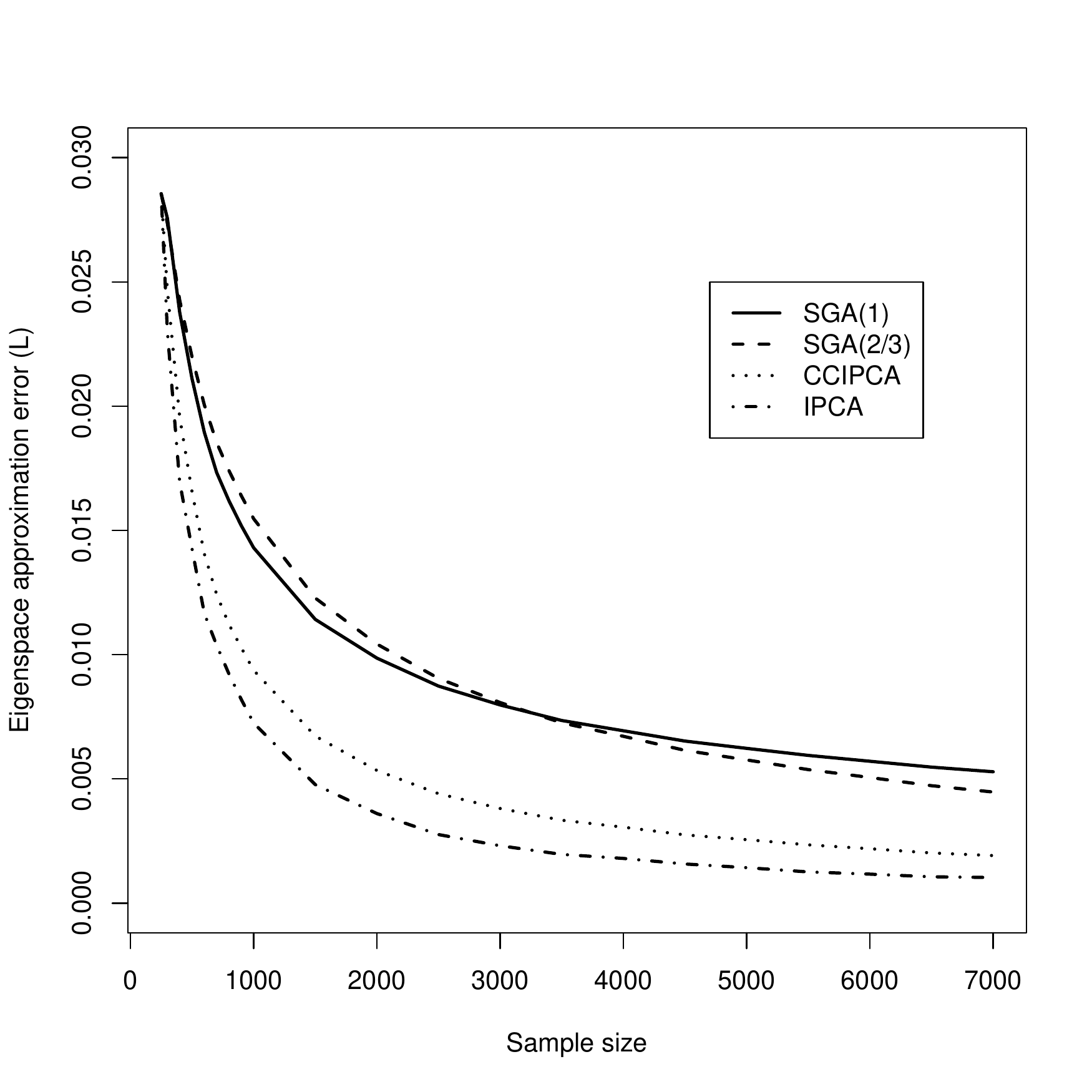}
 \caption{Eigenspace estimation error $L$ for the first $q=5$ eigenvectors of $\boldsymbol{\Gamma}$ with $d=100$}
% For SGA, the learning rate is of the form $c/n^\alpha$ with $\alpha\in\{ 1,2/3\}$. }
 \label{fig:errvecp100}
 \end{center}
   \end{figure}

\begin{figure}[h]
   \begin{center}
  \includegraphics[width=14cm]{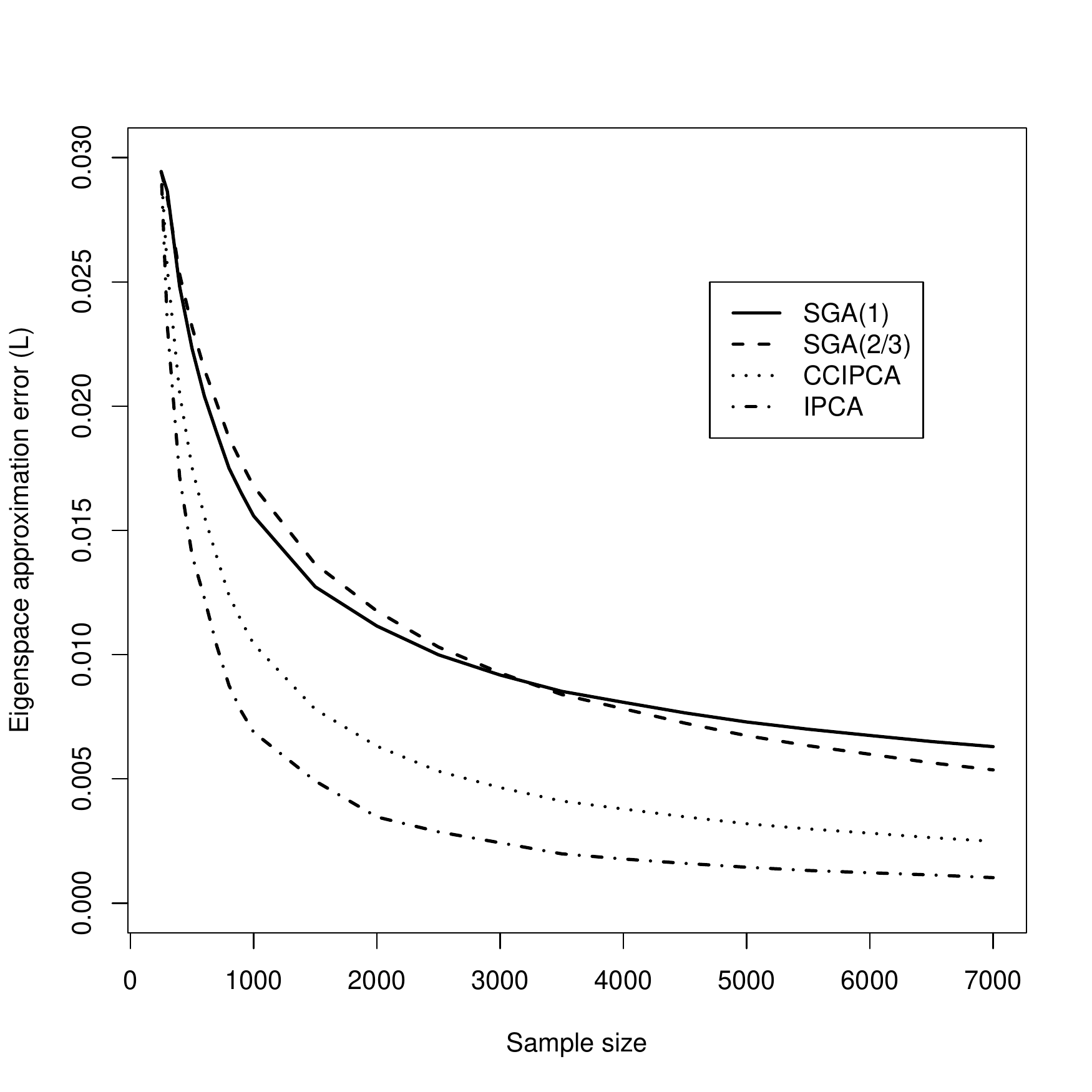}
 \caption{Eigenspace estimation error $L$ for the first $q=5$  eigenvectors of $\boldsymbol{\Gamma}$ with $d=1000$}
% For SGA, the learning rate is of the form $c/n^\alpha$ with $\alpha\in\{ 1,2/3\}$.}
 \label{fig:errvecp1000}
 \end{center}
   \end{figure}

Figures \ref{fig:errvecp100}--\ref{fig:errvecp1000} present the eigenspace estimation error $L$ (averaged over 100 replications) in function of the sample size $n $ for the most effective online algorithms under study: IPCA, SGA, and CCIPCA. Although the data dimension is $d=100$ in Figure \ref{fig:errvecp100} and  $d=1000$ in Figure \ref{fig:errvecp1000}, the two figures are similar, meaning that the effect of the dimension $d$ on the evolution of the accuracy is not crucial. IPCA produces reliable estimates and always outperform the SGA and CCIPCA algorithms. 
The SGA algorithm with learning rate $\gamma_n=c/n$ produces a stronger  initial decrease in $L$  than with  $\gamma_n = c/n^{2/3}$. In the long run however, $L$ decreases faster with the slower rate $ n^{-2/3}$.  
%because allows the algorithm to keep learning long after  $\gamma_n = c/n$ has converged to zero.  

\begin{table}[htdp]
\caption{Eigenspace estimation error $L$ for the first $q=5$ eigenvectors of $\boldsymbol{\Gamma}$} 
% The online algorithms are initialized with the batch PCA of the first $n_0=250$ observations. }
%latex.default(round(t(Table3$mean), 3), file = "table3m.tex",     title = "", rowname = c("Batch (n0)", "Bathc (n)", "SGA (1)",         "SGA (2/3)", "GHA (1)", "GHA (2/3)", "CCIPCA", "Incremental PCA",         "Perturbation", "Secular"), cgroup = c("$n=500$", "$n=1000$"),     n.cgroup = c(3, 3), cgroupTexCmd = "", colheads = c("$d=10$",         "$d=100$", "$d=1000$", "$d=10$", "$d=100$", "$d=1000$"))%
\begin{center}
\begin{tabular}{lrrrcrrr}
\hline\hline
\multicolumn{1}{l}{\ }&\multicolumn{3}{c}{\ $n=500$}&\multicolumn{1}{c}{\ }&\multicolumn{3}{c}{\ $n=1000$}\tabularnewline
\cline{2-4} \cline{6-8}
\multicolumn{1}{l}{}&\multicolumn{1}{c}{$d=10$}&\multicolumn{1}{c}{$d=100$}&\multicolumn{1}{c}{$d=1000$}&\multicolumn{1}{c}{}&\multicolumn{1}{c}{$d=10$}&\multicolumn{1}{c}{$d=100$}&\multicolumn{1}{c}{$d=1000$}\tabularnewline
\hline
Batch ($n_0$)&$0.041$&$0.027$&$0.032$&&$0.041$&$0.028$&$0.031$\tabularnewline
Batch ($n$)&$0.020$&$0.014$&$0.014$&&$0.010$&$0.007$&$0.007$\tabularnewline
SGA ($\alpha=1$)&$0.031$&$0.020$&$0.021$&&$0.025$&$0.014$&$0.016$\tabularnewline
SGA ($\alpha=2/3$)&$0.033$&$0.021 $&$ 0.023$&&$ 0.026$&$ 0.015$&$ 0.017$\tabularnewline
GHA ($\alpha=1$)&$0.030$&$0.020$&$0.023$&&$0.024$&$0.014$&$0.016$\tabularnewline
GHA ($\alpha=2/3$)&$0.032$&$0.021$&$0.023$&&$0.026$&$0.015$&$0.017$\tabularnewline
CCIPCA&$0.026$&$0.016$&$0.016$&&$0.016$&$0.010$&$0.010$\tabularnewline
IPCA&$0.020$&$0.015$&$0.015$&&$0.011$&$0.007$&$0.007$\tabularnewline
Perturbation&$0.546$&$1.697$& $1.997$ &&$0.499$&$1.727$& $1.989$ \tabularnewline
Secular&$0.020$&$0.014$&$0.014$ &&$0.010$&$0.007$& $0.007$ \tabularnewline
\hline
\end{tabular}\end{center}
\label{tab:errvecp}
\end{table}%

The random vector $\mathbf{X}$ considered in our simulation framework can be seen as a discretized standard Brownian motion. Given the smoothness of the eigenfunctions associated to the Brownian motion (sine functions) and the high dimension $d$ of the data, the functional PCA approach of Section \ref{fpca} seems an excellent candidate for estimating the eigenelements of the covariance matrix $\boldsymbol{\Gamma}$.  
To implement this approach, we  used a basis of $p=28$ cubic B-splines controlled by equispaced knots. 
%is continuous trajectories of a continuous time stochastic process, a competitive way to reduce both the storage of the data and the computation time consists in projecting the original data  onto a small number $p \ll d$ of basis functions. In our simulation framework, each $\mathbf{X}_i$ can be seen as a discretized sample path of a Brownian motion and  we have used 
The penalty matrix $\mathbf{P}$ was as in Section \ref{fpca}, namely, a roughness penalty on the second derivative 
  of the approximating function, with a smoothing parameter $\alpha = 10^{-7}$.
%  (see \cite{OSullivan86}) in order to avoid well known edge effects.
 The estimation error $L$ is presented in Table~\ref{tab:errvecpfda} for $d=100$ and $d=1000$. 
 Thanks to the dimension reduction, all online PCA algorithms can be rapidly computed even when $d=1000$. Comparing Tables \ref{tab:errvecp} and \ref{tab:errvecpfda}, it is clear that FPCA %smoothing the original discretized trajectories 
 improves the statistical accuracy. %estimation of eigenvectors. 
Again, the performances of the stochastic approximation approaches strongly depend on the dimension $d$ and the learning rate $\gamma_n$. As before, IPCA offers a good compromise between computation time and accuracy.

\begin{table}[h!]
\caption{Eigenspace estimation error $L$ for the first $q=5$  eigenvectors of $\boldsymbol{\Gamma}$ 
when the discretized trajectories are approximated by  spline functions with 28 equispaced knots}
\begin{center}
\begin{tabular}{lrrcrr}
\hline\hline
\multicolumn{1}{l}{\ }&\multicolumn{2}{c}{\ $n=500$}&\multicolumn{1}{c}{\ }&\multicolumn{2}{c}{\ $n=1000$}\tabularnewline
\cline{2-3} \cline{5-6}
\multicolumn{1}{l}{}&\multicolumn{1}{c}{$d=100$}&\multicolumn{1}{c}{$d=1000$}&\multicolumn{1}{c}{}&\multicolumn{1}{c}{$d=100$}&\multicolumn{1}{c}{$d=1000$}\tabularnewline
\hline
Batch ($n_0$)&$0.0244$&$0.0245$&&$0.0251$&$0.0227$\tabularnewline
Batch ($n$)&$0.0120$&$0.0114$&&$0.0060$&$0.0055$\tabularnewline
SGA ($\alpha=1$)&$0.0241$&$0.0242$&&$0.0245$&$0.0221$\tabularnewline
SGA ($\alpha=2/3$)&$0.0242$&$0.0242$&&$0.0246$&$0.0222$\tabularnewline
GHA ($\alpha=1$)&$0.0241$&$0.0242$&&$0.0245$&$0.0221$\tabularnewline
GHA ($\alpha=2/3$)&$0.0242$&$0.0242$&&$0.0246$&$0.0222$\tabularnewline
CCIPCA&$0.0149$&$0.0148$&&$0.0090$&$0.0080$\tabularnewline
IPCA&$0.0120$&$0.0114$&&$0.0060$&$0.0055$\tabularnewline
Perturbation&$0.6085$&$0.6137$&&$0.6249$&$0.5962$\tabularnewline
Secular&$0.0120$&$0.0114$&&$0.0060$&$0.0055$\tabularnewline
\hline
\end{tabular}
\end{center}
\label{tab:errvecpfda}
\end{table}

\begin{comment}
 \begin{table}[hpt]
\caption{Mean estimation errors ($\times 100$) for the eigenspace generated by the first eigenvector ($q=1$) and different values of $d$ and  sample of size $n$. The incremental algorithms starts once a sample of size $n_0=250$ has been used to compute the initialization eigenvectors. (In parentheses the median errors).}
{\small
\begin{center}
\begin{tabular}{|ccccccc|} \hline
 & \multicolumn{3}{c}{$n=500$} & \multicolumn{3}{c|}{$n=1000$} \\
&  $d=10$ & $d=100$ &  $d=200$ & $d=10$ & $d=100$ & $d=200$ \\ \hline
Batch PCA ($n_0$) & 0.215 (0.176) & 0.231 (0.176) &  0.211 (0.168) &  0.205 (0.145) & 0.222 (0.164) &   0.229 (0.175) \\
Batch PCA ($n$)  &0.106 (0.084) &   0.120 (0.094) & 0.105 (0.083)   &0.050 (0.038) &   0.055 (0.041) & 0.054 (0.042) \\ \hline
SGA ($\alpha=1$)  &0.336 (0.290) &  1.85 (1.65)  & 3.52 (3.04) &  0.215 (0.166) &  1.08 (0.943) & 2.14 (1.78) \\
SGA ($\alpha=2/3$)  &0.965 (0.853) &  6.39 (5.55) & 13.3 (11.0)   &0.647 (0.548) &   5.380 (4.84) & 13.9 (10.9) \\ 
Averaging & 0.170 (0.120) & 0.270 (0.287) &  0.558 (0.287) & 0.134 (0.072) &  0.609 (0.331) &    2.27 (1.14) \\ \hline
Rank $1$ inc. &  0.106 (0.083)   &  0.120 (0.094) &  0.105 (0.083) &   0.038 (0.026) &  0.055 (0.041) & 0.054  (0.042) \\ \hline
\end{tabular}
\end{center}
}
\label{tab:errvecpav}
\end{table}
\end{comment}

\subsubsection{Missing data}

\begin{figure}[h!]
   \begin{center}
  \includegraphics[height=14cm,width=15.5cm]{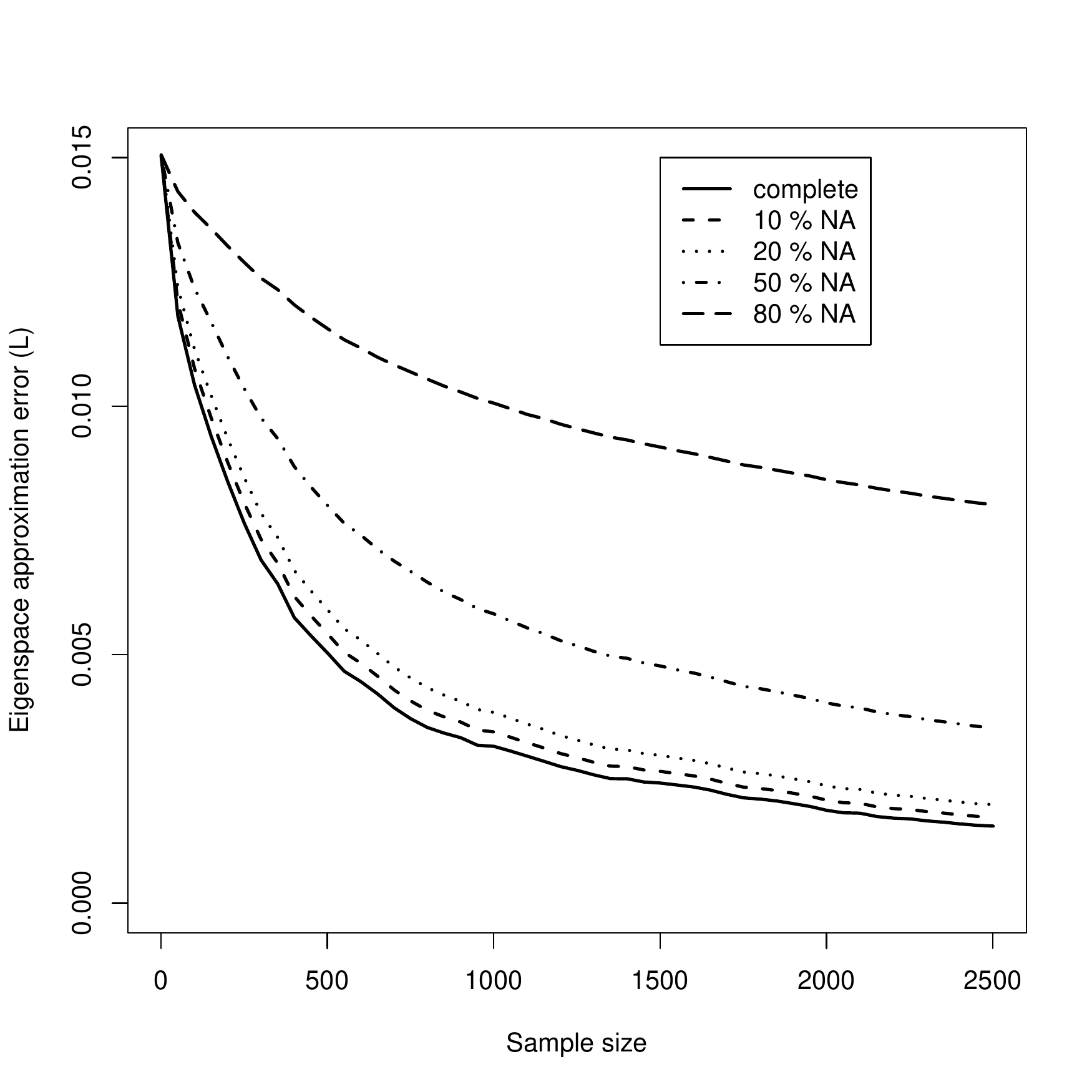}
 \caption{Eigenspace estimation error $L$ of the IPCA algorithm for the first $q=2$ eigenvectors 
 of $\boldsymbol{\Gamma}$ with different levels of missingness in the data and $d=1000$}
 %in function of the sample size $n$. D
 %are considered  (0\%, 10\%, 20\%, 50\%, and 80\%) }
 \label{fig:errNA1000}
 \end{center}
   \end{figure}

The ability of the IPCA algorithm to handle missing data was evaluated in a high dimensional context ($d=1000$) using the EBLUP imputation method of Section \ref{sec: missing data}. % (\ref{def:imputincRpca}).  
Missing values were removed by simple random sampling without replacement with different sampling fractions ($f=0, 0.1, 0.2, 0.5, 0.8$). As shown in Figure~\ref{fig:errNA1000}, the incremental algorithm performs well even when half of the data are missing. We note  that imputation based on EBLUP is well adapted to this simulation study 
due to the rather strong correlation between variables. It is also worth noting that the computation time if the imputation is about of the same order as the computation time of the IPCA algorithm itself.

\subsection{A face recognition example}

To assess the performance of online PCA with real data, we have selected the Database of Faces of the  AT\&T Laboratories Cambridge (\url{http://www.cl.cam.ac.uk/research/dtg/attarchive/facedatabase.html}). This database consists in 400 face images of dimensions $92 \times 112$  pixels in 256 gray levels. For each of 40 subjects, 10 different images featuring various facial expressions (open/closed eyes, smiling/non-smiling) and facial details (glasses/no-glasses) are available.

%applied the most effective algorithms previously found  to 

The database was randomly split in a training set and a test set by stratified sampling. Specifically, for each subject, one  image was randomly selected for testing and the nine others were included in the training set. The IPCA, SGA, and CCIPCA algorithms were applied to the (vectorized) training images using $q=20$ or $q=40$ principal components as in \cite{LevyLindenbaum2000}. No image centering was used (uncentered PCA). IPCA and CCIPCA were initialized using only the first image whereas the SGA algorithm was initialized with the batch PCA of the first $q$ images. 
The resulting principal components were used for two tasks: compression and classification of the test images. 
We also computed the batch PCA of the data as a benchmark for the online algorithms. 

For the compression task, we measured the performance of the algorithms
using the uncentered, normalized version of the loss function \eqref{def:riskPCA}: 
\begin{equation*}
R_n(\widehat{\mathbf{P}}_q) =  \frac{1}{n}  \sum_{i=1}^n \frac{ \big\|  \mathbf{x}_i    - \widehat{\mathbf{P}}_q  \mathbf{x}_i   \big\|^2 }{   \left\|  \mathbf{x}_i    \right\|^2 }\,  .
\end{equation*}

For the purpose of classifying the test images, we performed a linear discriminant analysis (LDA) of the 
scores of the training images on the principal components, using the subjects as classes. This technique is a variant of the well-known Fisherface method of \cite{BHK1997}; %(In the Fisherface method, the PCA step computes $q=d-c$ principal components, with $d$ the image dimension and $c$ the number of classes.) 
see also \cite{Zhao2006} for related work.

\begin{figure}[h!]
\begin{center}
\begin{tabular}{ccccc}
\includegraphics[width=0.18\textwidth]{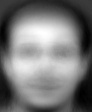} &
\includegraphics[width=0.18\textwidth]{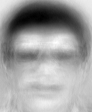} &
\includegraphics[width=0.18\textwidth]{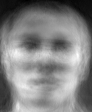} &
\includegraphics[width=0.18\textwidth]{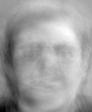} &
\includegraphics[width=0.18\textwidth]{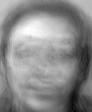} \\
\includegraphics[width=0.18\linewidth]{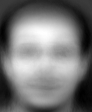} &
\includegraphics[width=0.18\linewidth]{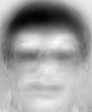} &
\includegraphics[width=0.18\linewidth]{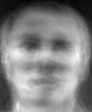} &
\includegraphics[width=0.18\linewidth]{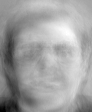} &
\includegraphics[width=0.18\linewidth]{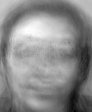} \\
\includegraphics[width=0.18\linewidth]{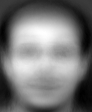} &
\includegraphics[width=0.18\linewidth]{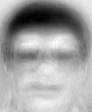} &
\includegraphics[width=0.18\linewidth]{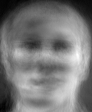} &
\includegraphics[width=0.18\linewidth]{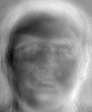} &
\includegraphics[width=0.18\linewidth]{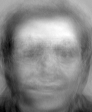} \\
\includegraphics[width=0.18\linewidth]{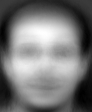} &
\includegraphics[width=0.18\linewidth]{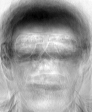} &
\includegraphics[width=0.18\linewidth]{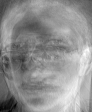} &
\includegraphics[width=0.18\linewidth]{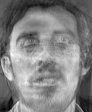} &
\includegraphics[width=0.18\linewidth]{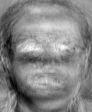} 
\end{tabular}
\end{center}
\caption{Top five principal components (eigenfaces) for the AT\&T Database of Faces. 
Rows from top to bottom: batch PCA, IPCA, CCIPCA, SGA}
\label{fig:eigenfaces}
\end{figure}

\begin{figure}[h!]
\begin{center}
\begin{tabular}{cccccc}
\includegraphics[width=0.14\textwidth]{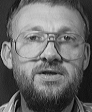} &
\includegraphics[width=0.14\textwidth]{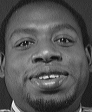} &
\includegraphics[width=0.14\textwidth]{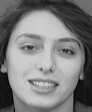} &
\includegraphics[width=0.14\textwidth]{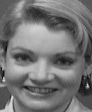} &
\includegraphics[width=0.14\textwidth]{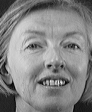} &
\includegraphics[width=0.14\textwidth]{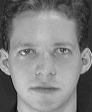} \\
\includegraphics[width=0.14\textwidth]{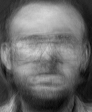} &
\includegraphics[width=0.14\textwidth]{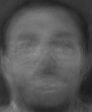} &
\includegraphics[width=0.14\textwidth]{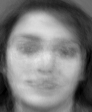} &
\includegraphics[width=0.14\textwidth]{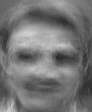} &
\includegraphics[width=0.14\textwidth]{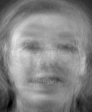} &
\includegraphics[width=0.14\textwidth]{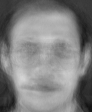} \\
\includegraphics[width=0.14\linewidth]{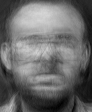} &
\includegraphics[width=0.14\linewidth]{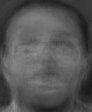} &
\includegraphics[width=0.14\linewidth]{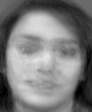} &
\includegraphics[width=0.14\linewidth]{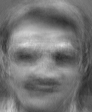} &
\includegraphics[width=0.14\linewidth]{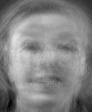} &
\includegraphics[width=0.14\linewidth]{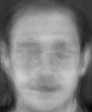} \\
\includegraphics[width=0.14\linewidth]{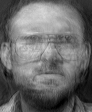} &
\includegraphics[width=0.14\linewidth]{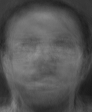} &
\includegraphics[width=0.14\linewidth]{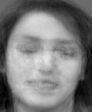} &
\includegraphics[width=0.14\linewidth]{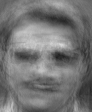} &
\includegraphics[width=0.14\linewidth]{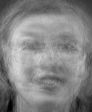} &
\includegraphics[width=0.14\linewidth]{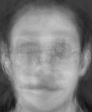} \\
\includegraphics[width=0.14\linewidth]{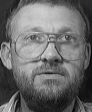} &
\includegraphics[width=0.14\linewidth]{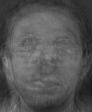} &
\includegraphics[width=0.14\linewidth]{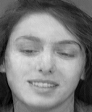} &
\includegraphics[width=0.14\linewidth]{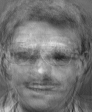} &
\includegraphics[width=0.14\linewidth]{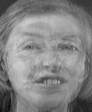} &
\includegraphics[width=0.14\linewidth]{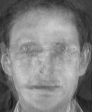} 
\end{tabular}
\end{center}
\caption{Sample of compressed images from the AT\&T Database of Faces ($q=40$ principal components). 
Rows from top to bottom: original image, batch PCA, IPCA, CCIPCA, SGA}
\label{fig:compressed images}
\end{figure}

The random split of the data was repeated 100 times for each task and we report here the average results. Table \ref{face compression} reports the performance of the algorithms with respect to data compression. As can be expected, the compression obtained with $q=40$ principal components is far superior to the one using $q=20$ components. Due to the large size of the training set (90\% of all data), there is little difference in compression error between the training and testing sets. IPCA and CCIPCA produce nearly optimal results that are almost identical to batch PCA (see also Figures ref). The SGA algorithm shows worse performance on average but also more variability. The effectiveness of SGA further degrades if a random initialization is used. Interestingly, some principal components found by SGA strongly differ from those of the other methods, as shown in Figure \ref{fig:eigenfaces} where SGA components are stored in the last row. Accordingly, SGA may compress images quite differently from the other methods. In Figure \ref{fig:compressed images} for example, some images compressed  by SGA (first and third images in the last row) have sharp focus and are in fact very similar to other images of the same subjects used in the training phase. In contrast, the corresponding images compressed by batch PCA, IPCA, and CCIPCA are blurrier, yet often closer to the original image.

\begin{table}[htdp]
\caption{Compression loss of the batch PCA, IPCA, CCIPCA, and SGA algorithms with the AT\&T Database of Faces}
\begin{center}
\begin{tabular}{l c c c c c}
\hline 
\hline
& \multicolumn{2}{c}{$q=20$} & \multicolumn{1}{c}{ \ } & \multicolumn{2}{c}{$q=40$} \\ 
\cline{2-3} \cline{5-6}
& Training & Test & & Training & Test \\
\hline
Batch & 0.0323 &0.0363&& 0.0224 & 0.0286 \\
IPCA &0.0327 &0.0367&&0.0229 & 0.0290 \\
SGA   &0.0523 &0.0551&&0.0388  & 0.0431 \\
CCIPCA  &0.0335& 0.0373&&0.0257 & 0.0312 \\
\hline
\end{tabular}
\end{center}
\label{face compression}
\end{table}%

Table \ref{face classification} displays the classification accuracy of the LDA  based on the component scores of the different online PCA algorithms. Overall, all algorithms have high accuracy. IPCA and CCIPCA yields the best performances, followed closely by batch PCA. It is not surprising that online algorithms can surpass batch PCA in classification since the latter technique is only optimal for data compression. SGA produces slightly lower, yet still high classification accuracy.

\begin{table}[htdp]
\caption{Classification rates for batch PCA, IPCA, CCIPCA, and SGA coupled with Linear Discriminant Analysis on the AT\&T Database of Faces}
\begin{center}
\begin{tabular}{l c c c c c}
\hline 
\hline
& \multicolumn{2}{c}{$q=20$} & \multicolumn{1}{c}{ \ } & \multicolumn{2}{c}{$q=40$} \\ 
\cline{2-3} \cline{5-6}
& Training & Test & & Training & Test \\
\hline
Batch & 0.9897  &  0.9580 &&0.9986 & 0.9880 \\
IPCA & 0.9915   & 0.9635 && 0.9995&  0.9875 \\
CCIPCA & 0.9920 &   0.9655&&0.9988 &  0.9837 \\
SGA &  0.9788 & 0.9340 &&   0.9963 &  0.9710 \\
\hline
\end{tabular}
\end{center}
\label{face classification}
\end{table}%

Table \ref{table:face space-time usage} examines the computation time and memory usage of the PCA algorithms. 
As can be expected, batch PCA requires much more (at least one order of magnitude) memory than the online algorithms. Also, the size of the data is large enough so that batch PCA becomes slower than the online algorithms CCIPCA and IPCA. The fact that the SGA algorithm runs slower than all other algorithms 
 is not surprising given that its exact implementation used here requires a 
 Gram-Schmidt orthogonalization in high dimension at each iteration.

\begin{table}[h!]
\caption{Computation time (s) and memory usage (MB) for the PCA of the AT\&T Database of Faces ($n=400$, $d=10304$). 
}
\begin{center}
\begin{tabular}{l  c c}
\hline
Method & Time  & Memory  \\
\hline
Batch PCA & 7.13 & 924.4 \\
IPCA & 7.09 & 73.3 \\
CCIPCA & 3.93 & 74.6 \\
SGA & 9.45 &  67.8 \\
\hline 
\end{tabular}
\end{center}
\label{table:face space-time usage}
\end{table}%

%Empirical compression loss for $k=1,\ldots, n$:
%\begin{equation*}
%\mathrm{Comp}_k (\widehat{\mathbf{P}}_{q,k} ) = \frac{ \sum_{i=1}^k \left\| \big(\mathbf{I}_d - \widehat{\mathbf{P}}_{q,k}\big)
%\left(\mathbf{X}_i -\boldsymbol{\mu}_k \right) \right\|^2 }{  \sum_{i=1}^k \left\| \mathbf{X}_i -\boldsymbol{\mu}_k \right\|^2}
%\end{equation*}
%
%Error in the estimation of the (linear space generated by the) first $q$ eigenvectors:
%\begin{equation*}
%R (\widehat{\mathbf{P}}_{q,k} ) = \mathrm{tr} \left[ \left( \widehat{\mathbf{P}}_{q,k} - \mathbf{P}_q \right)^T \left( \widehat{\mathbf{P}}_{q,k} - \mathbf{P}_q \right) \right] /  
%\end{equation*}
%

%%%%%%%%%%%%%%%%%%%%%%%%%%%%%%%%%%%%%%%%%%%%%%%%%%%%%%%%%%%%%%%

\section{Concluding remarks}
\label{sec: conclusion}

PCA is a popular and powerful tool  for the analysis of high-dimensional data with multiple applications to data mining, data compression, feature extraction, pattern detection, process monitoring, fault detection, and computer vision. 
We have presented several online algorithms that can efficiently perform and update the PCA of time-varying data (e.g., databases, streaming data) and massive datasets. We have compared the computational and statistical performances of these algorithms using artificial and real data. The R package \texttt{onlinePCA} available at \url{http://cran.r-project.org/package=onlinePCA} implements all the techniques discussed in this paper and others.   

Of all  algorithms under study, the stochastic methods SGA, SNL, and GHA provide the highest computation speed. They are however very sensitive to the choice of the learning rate (or step size) and converge more slowly than IPCA and CCIPCA. For strongly misspecified learning rates, they may even fail to converge. 
%produce highly inaccurate estimates. 
%, even worse than PCA performed on a small subsample of the data. 
Furthermore, simulations not presented here suggest that a different step size should be used for each estimated eigenvector. In theory this guarantees the almost sure convergence of estimators towards the corresponding  eigenspaces  \citep{Monnez} but, as far as we know,  there exist no automatic procedure for choosing these $q$ learning rates in practice. 
In relation to the recent result given in \cite{BalsubramaniDF2013}, averaging techniques (see \cite{PolyakJud92}) could be useful to get efficient estimators of the first eigenvector. Simulation studies not presented here do not confirm at all this intuition. As a matter of fact, averaging improves significantly the initial stochastic gradient estimators when $0.5<\alpha<1$ but the estimation error remains much larger than with IPCA  and CCIPCA.
 
The IPCA and CCIPCA algorithms offer a very good compromise between statistical accuracy and computational speed. 
They also have the advantage of not having major dependence on tuning parameters (forgetting factor).  
Approximate perturbation methods can yield highly inaccurate estimates and we do not recommend them in practice. The method of secular equations, although slower than the other algorithms, has the advantage of being exact. It is a very good option when accuracy matters more than speed and the dimension $d$ is not too large. In particular, it is very effective with functional data that have been projected onto a small number of basis functions (FPCA). More generally, when applicable, FPCA should be preferred over standard PCA as it demonstrates both higher accuracy and higher computation speed.   
% If only one iteration (one update)  is needed,   
%In the case of functional data, associated with reduction dimension based on  it really becomes interesting. 
In the presence of missing data, imputation procedures like the EBLUP of \cite{Brand02incrementalsingular} enable online PCA algorithms to continue running without considerable increase in computation time or decrease in accuracy. 

For reasons of space, we have focused on rank-1 PCA updates in this paper. However block updates of rank $r \ge 2$ are also frequent in practice. The user choice of the block size $r$ has complex effects on the accuracy and speed of algorithms: for instance, larger blocks tend to reduce noise and estimation variability, but they may also slow down convergence. Regarding computations, as $r$ increases the running time initially decreases but then it reaches a plateau  and may even increase if $r$ is too large. 
%A rule of thumb is to take $r$ of the same order as the number $q$ of eigenvectors to compute \citep[e.g.,][]{LevyLindenbaum2000}. 
In additional simulations (see Supplementary Materials) we examined two online PCA algorithms that allow for block updates: the IPCA algorithm of \cite{LevyLindenbaum2000} and the block-wise stochastic power method of  \cite{Mitliagkasetal2014}. 
In the former algorithm, a rule of thumb  is to take $r$ of the same order as the number $q$ of eigenvectors to compute. 
With the choice $r=q$, this algorithm was actually faster than the fast implementations of the SGA, SNL, and GHA while maintaining the very high accuracy of the rank-1 update IPCA of Section \ref{sec:incremental}. 
The block-wise stochastic power method was even much faster and, using the recommended block size $r \approx \log(d)/n $, as accurate as the stochastic algorithms. 

%Finally, we point out that for the analysis of static datasets of moderate to large size, say $\min(n,d)\le 10^4$, 

%\begin{tabular}{ |p{3cm}||p{3cm}|p{3cm}|p{3cm}|  }

%\begin{table}
%\begin{sideways}
%\begin{tabular}{ p{2.2cm}   c  c  p{1.8cm}  p{2.8cm}  c c c p{3cm} }
%%\begin{tabular}{ c c c c c  }
%\hline
%Method  & Computation & PCA rank & Tuning \ parameter & Orthonormality & Update & Runtime & Space & Additional comments \\
%\hline
%Perturbation &  Approximate & Full & No & No & single & $O(d^2)$ & $O(d^2)$ & Only accurate for small $d$ after many iterations  \\
%Secular equations  & Exact & Full & No & No  & single & & $O(d^2)$ & Loss of orthonormality for large $d$ and/or many iterations  \\
%Secular equations-R & Approximate & Full & No & Yes  & single & & $O(d^2)$ &  \\
%IPCA  & Approximate & Reduced & No & No   \\
%IPCA-R  & Approximate & Reduced & No & Yes   \\
%SGA-R   & N/A & Reduced & Yes & Yes   \\
%SGA-NN   & N/A &   Reduced & Yes  & Yes   \\
%SNL-R   & N/A &   Reduced & Yes  & Yes  \\
%SNL-NN   & N/A  & Reduced & Yes  & Yes \\
%\hline
%\end{tabular}
%\end{sideways}
%\caption{test}
%\end{table}
%

%%%%%%%%%%%%%%%%%%%%%%%%%%%%%%%%%%%%%%%%%%%%%%%%%%%%%%%%%%%%%%%

%%%%% Biblio 
\bibliographystyle{apalike}

\bibliography{OnlineFPCA}

\end{document}